\documentclass{article}
\usepackage[T1]{fontenc}
\usepackage{iclr2027_conference,times}
\usepackage{amsmath,amssymb,amsthm}
\usepackage{graphicx,booktabs,multirow,array}
\usepackage{colortbl}
\usepackage{caption}
\definecolor{pbrgray}{RGB}{240,240,240}
\definecolor{branchblue}{RGB}{228,240,249}
\usepackage{algorithm,algpseudocode}
\usepackage{placeins}
\usepackage{float,needspace}
\usepackage{hyperref,url}
\newtheorem{proposition}{Proposition}
\newcommand{\E}{\mathbb{E}}
\newcommand{\Var}{\operatorname{Var}}
\newcommand{\KL}{\operatorname{KL}}
\newcommand{\sg}{\operatorname{sg}}
\newcommand{\clip}{\operatorname{clip}}
\newcount\tablecolumns
\makeatletter
\newcommand{\countTableColumns}[1]{\tablecolumns=0\@tfor\tablecolumn:=#1\do{\advance\tablecolumns by 1}}
\makeatother
\newcommand{\fitTable}[1]{\par\noindent\makebox[\linewidth][c]{\begingroup\setbox0=\hbox{#1}\ifdim\wd0>\linewidth\resizebox{\linewidth}{!}{\box0}\else\let\originaltabular\tabular\renewcommand{\tabular}[1]{\countTableColumns{##1}\setlength{\tabcolsep}{\dimexpr\tabcolsep+(\linewidth-\wd0)/(2*\tablecolumns)\relax}\originaltabular{##1}}#1\fi\endgroup}\par}
\newcommand{\method}{BranchOPD}

\newcommand{\homepage}{\raisebox{-1.6pt}{\includegraphics[height=1.05em]{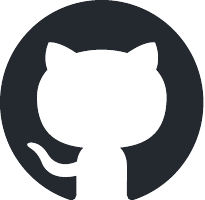}}}
\newcommand{\hfmodel}{\raisebox{-1.6pt}{\includegraphics[height=1.05em]{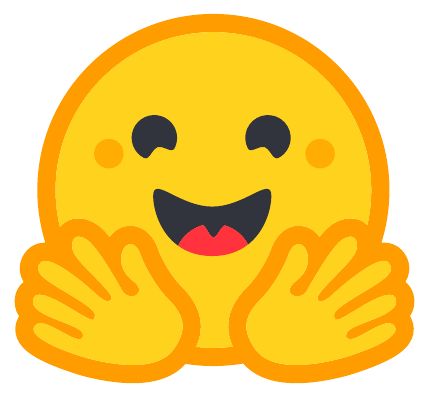}}}
\definecolor{titlelink}{RGB}{60,60,60}
\newcommand{\titlelink}[2]{{\hypersetup{urlcolor=titlelink}\href{#1}{#2}}}
\title{Branching On-Policy Distillation}
\author{%
\bfseries Dongyang Chen$^{1,3}$, Shudong Liu$^{2,3}$, Shubo Li$^{3}$, Duoxun Tang$^{1}$,\\
\bfseries Jinwei Liang$^{3}$, Zehuan Wu$^{3}$, Lewei Lu$^{3}$\thanks{Corresponding author.}\\[1mm]
\normalfont $^{1}$Tsinghua University\quad $^{2}$Peking University\quad $^{3}$SenseTime Research\\[2.5mm]
\normalfont\small
\homepage\,\titlelink{https://github.com/chendy25/BranchOPD.git}{github.com/chendy25/BranchOPD}%
\hspace{2.5em}%
\hfmodel\,\titlelink{https://huggingface.co/BranchOPD}{huggingface.co/BranchOPD}%
}
\iclrfinalcopy
\renewcommand{\headrulewidth}{0pt}
\hypersetup{pdftitle={Branching On-Policy Distillation},pdfauthor={Dongyang Chen, Shudong Liu, Shubo Li, Duoxun Tang, Jinwei Liang, Zehuan Wu, Lewei Lu},colorlinks=true,citecolor=blue,linkcolor=blue,urlcolor=blue}
\begin{document}
\maketitle
\begin{abstract}
On-policy distillation (OPD) provides teacher supervision on student-generated trajectories, but a single trajectory realizes only one continuation at each decision, limiting supervision coverage under a finite sampling budget. We propose \textbf{BranchOPD (Branching On-Policy Distillation)}, which selects informative decision points using student uncertainty and teacher--student disagreement, resamples student-policy continuations from shared prefixes, and applies token-level teacher supervision to both main trajectories and branches. BranchOPD operates at token-level decisions in mathematics and within-turn decisions in agent tasks. Shared-prefix reuse reduces student-generated token costs by approximately 19\% on ALFWorld and 31\% on WebShop relative to seven independent trajectories. We also introduce protocol-boundary repair (PBR), which uses targeted supervision under protocol constraints to mitigate repetition and wasted generation. BranchOPD outperforms Vanilla OPD and Multi-Rollout OPD ($n=7$) across all evaluated benchmarks, with gains over Vanilla OPD of 18.55 percentage points in mean mathematics accuracy over 16 sampled answers (avg@16) and 13.8 points in mean ALFWorld ID/OOD success. Mechanistic analyses and selection ablations show that branching at uncertain decisions expands subsequent behavior coverage and outperforms random selection and turn-initial branching.
\end{abstract}

\section{Introduction}
Distillation is a standard way to transfer the reasoning and interaction abilities of a large language model to a smaller one. Training on fixed teacher outputs \citep{ref19,ref20} exposes the student only to prefixes the teacher would write, whereas at inference the student must continue from its own. On-policy distillation (OPD) closes this gap: the student generates trajectories and the teacher scores every generated token \citep{ref1,ref21,ref22,ref23,ref24}.

\begin{figure}[t]
\centering
\includegraphics[width=\linewidth]{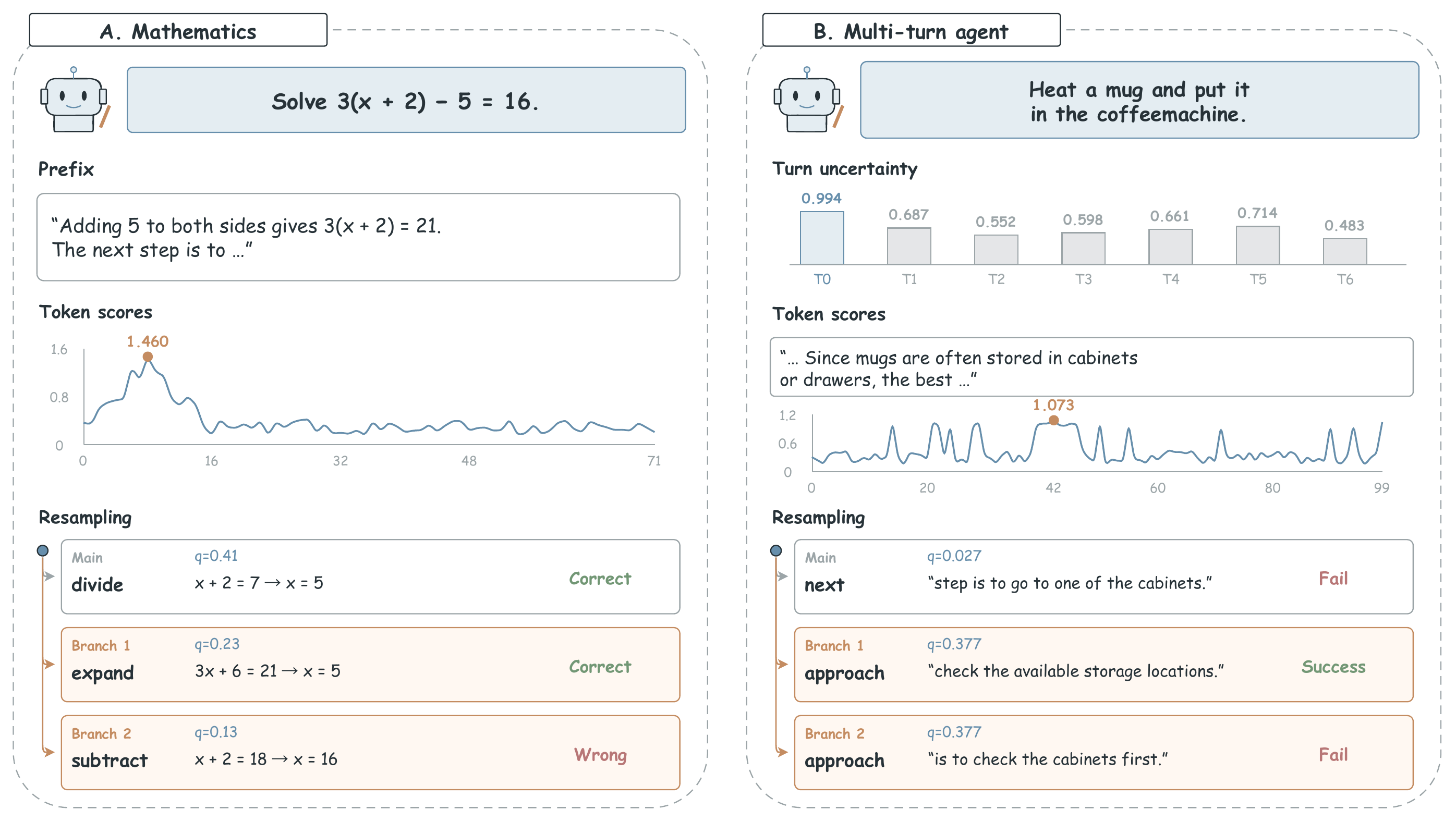}
\caption{Resampling from one prefix on a mathematics problem and a logged ALFWorld trajectory. Scores mark the fork; $q$ is the student probability of each continuation's first token.}
\label{fig:intro}
\end{figure}

Each trajectory, however, follows only one of the continuations available at every decision. Even with the teacher's full next-token distribution at a prefix, the student receives no supervision on the multi-step histories that the unchosen tokens would have led to. This matters in mathematical reasoning, where an early step constrains the rest of the derivation, and in agent tasks, where an action changes all later observations \citep{ref25,ref26,ref30,ref31,ref56}. Sampling more independent trajectories widens coverage, but most of the extra tokens regenerate identical or near-identical prefixes. As Figure~\ref{fig:intro} shows, resampling from a single prefix can reach quite different paths, sometimes even when the first resampled token matches the original.

Recent OPD methods decide where to spend supervision on trajectories that have already been sampled: they reweight tokens, reasoning steps, or turns \citep{ref2,ref3,ref4,ref71}, order trajectories by difficulty \citep{ref18}, or reuse teacher-scored samples \citep{ref72}. We ask instead which \emph{additional} trajectories to sample. Tree search and branching reinforcement learning already share prefixes to explore locally \citep{ref8,ref9,ref10,ref11,ref27,ref28,ref29,ref68,ref69,ref70}; we bring this idea to distillation, where every branch token receives dense teacher supervision and no task reward is needed.

We propose \method{}. For each input, the student first generates a main trajectory. We then pick a few fork points using student uncertainty and teacher--student disagreement, and sample new continuations from the frozen student policy at each fork. In mathematics tasks, we select token positions directly; in agent tasks we first select turns and then positions within them. The teacher scores all generated tokens on the main trajectory and the branches, and a shared prefix enters the loss only once. Because branches come from the student rather than from the teacher or a refined trajectory \citep{ref6,ref73}, they stay on the student's own distribution.

A commonly encountered failure mode in OPD is improper termination: the student may continue generating after producing the final answer, often emitting redundant content or stray formatting tokens \citep{yang2026eos, luo2026demystifying}. We handle this problem by \emph{protocol-boundary repair} (PBR), an auxiliary loss on detected protocol events, and apply PBR to the baselines so that comparisons isolate the effect of branching.

Our contributions are:
\begin{itemize}
\item \textbf{Branching on-policy distillation.} A shared-prefix sampling scheme that selects informative decisions and resamples student continuations from them for token-level teacher supervision, in both mathematical reasoning and multi-turn agent tasks.
\item \textbf{Protocol-boundary repair.} An auxiliary loss that reinforces beneficial and suppresses harmful protocol events in student outputs, reducing repetition and wasted generation.
\item \textbf{Experiments and analysis.} \method{} outperforms Vanilla OPD and Multi-Rollout OPD on all evaluated benchmarks, including when every method uses PBR, while generating about 19\% and 31\% fewer student tokens than seven independent trajectories on ALFWorld and WebShop. Ablations and diagnostics show that the choice of fork points matters.
\end{itemize}

\section{Related Work}
\paragraph{On-policy distillation.}
GKD, MiniLLM, and DistiLLM train students on their own samples and study which divergence to minimize \citep{ref1,ref23,ref24}. Reasoning supervision can also come from solution selection, process feedback, or explanations \citep{ref35,ref36,ref37}. More recent OPD methods reweight supervision across tokens \citep{ref2,ref71}, reasoning steps \citep{ref3}, or interaction turns \citep{ref4}, or schedule trajectories by difficulty \citep{ref18}. EOPD adds a forward-KL term at tokens where the teacher has high entropy \citep{ref58}, and other work reformulates the target or constrains the update \citep{ref59,ref60}. Closest to ours, SPOT forces teacher-proposed tokens at selected positions, scores the resulting student continuations with an outcome verifier, and distills a calibrated local target \citep{ref79}. In \method{}, branches are sampled from the student itself and supervised token by token, without task rewards.

\paragraph{Quality of distillation supervision.}
Probability mismatch and degenerate agreement motivate studies of what makes OPD supervision useful \citep{ref5,ref7,ref17}. Trajectory refinement and expert corrections improve the examples shown to the student \citep{ref6,ref67}, and teacher continuations from student anchors extend supervision beyond sampled rollouts \citep{ref73}; our continuations come from the student. Our objective is token-local and complements work that separates token-level updates from sequence-level credit assignment \citep{ref75}. PBR follows unlikelihood training \citep{ref12} in supervising specific output events.

\paragraph{Tree search and reinforcement learning.}
Local and tree-structured exploration reuses prefixes to sample alternative decisions \citep{ref8,ref9,ref10,ref11}. Recent methods branch at high-entropy tokens with sibling-return baselines \citep{ref70}, adapt rollout budgets \citep{ref68,ref69}, or roll back the environment and reflect \citep{ref76}. These methods compare branches through task rewards; \method{} uses shared-prefix exploration to collect more tokens for teacher supervision. Appendix~\ref{app:background} covers further learning and systems work.

\section{Preliminaries}
\label{sec:prelim}
\subsection{Student trajectories}
Let $x\sim\mathcal D$ be an input, $\mathcal V$ the vocabulary, $\pi_\theta$ the student policy, and $\pi_T$ a frozen teacher. We write $a_t\in\mathcal V$ for a generated token and $s_t$ for its prefix. In mathematical reasoning, a trajectory is a single response $\tau=(x,a_1,\ldots,a_{|\tau|})$ that ends at a stop token or the length limit, with $s_t=(x,a_{<t})$ and $\pi_\theta(\tau\mid x)=\prod_{t}\pi_\theta(a_t\mid s_t)$. In multi-turn agent tasks, following \citet{ref4}, the student interacts with an environment $\mathcal E$ for up to $M$ turns:
\begin{align}
\tau&=(x,y_1,o_1,\ldots,y_M,o_M),\nonumber\\
\pi_\theta(\tau\mid x)&=\prod_{t\in\mathcal T(\tau)}\pi_\theta(a_t\mid s_t)\prod_{m=1}^{M}P_{\mathcal E}(o_m\mid x,y_{\leq m},o_{<m}).
\label{eq:agenttraj}
\end{align}
Each response $y_m$ is a token sequence that may contain reasoning and an environment action, and $o_m$ is the resulting observation. $\mathcal T(\tau)$ indexes the student-generated tokens, whose prefixes $s_t$ include earlier responses and observations; observations are never supervised. Mathematics is the special case $M=1$ without observations.

For $t\in\mathcal T(\tau)$, let $\tau_{\geq t}$ denote the continuation from $s_t$, including later environment transitions, and $\pi_\theta(\tau_{\geq t}\mid s_t)$ its distribution. To resume from $s_t$ in an agent task, the environment must be restored to its state at $s_t$, by snapshot or action replay, so that the resumed continuation has the same distribution as an uninterrupted rollout.

\subsection{On-policy distillation}
OPD trains the student on its own trajectories under teacher supervision \citep{ref1,ref23,ref24}. Trajectories are sampled from $\pi_{\theta_{\rm old}}$, a frozen copy of the student; we sample at temperature one without truncation, so $\pi_{\theta_{\rm old}}$ is exactly the sampling distribution. OPD minimizes the reverse KL divergence on student-visited prefixes:
\begin{equation}
\mathcal L_{\rm OPD}(\theta)=\E_{x\sim\mathcal D,\,\tau\sim\pi_{\theta_{\rm old}}(\cdot\mid x)}\Big[\frac{1}{|\mathcal T(\tau)|}\sum_{t\in\mathcal T(\tau)}\KL\big(\pi_\theta(\cdot\mid s_t)\,\Vert\,\pi_T(\cdot\mid s_t)\big)\Big].
\label{eq:opd}
\end{equation}
In practice, the teacher scores only the sampled tokens, which gives the token-level distillation signal
\begin{equation}
\Delta\log p_t=\log\pi_T(a_t\mid s_t)-\log\pi_\theta(a_t\mid s_t).
\label{eq:dlogp}
\end{equation}
If the teacher's support covers the student's, $\E_{a_t\sim\pi_\theta(\cdot\mid s_t)}[-\Delta\log p_t]$ equals the reverse KL at $s_t$. $\Delta\log p_t$ is positive when the teacher assigns the sampled token higher probability than the student does. With the importance ratio $r_t(\theta)=\pi_\theta(a_t\mid s_t)/\pi_{\theta_{\rm old}}(a_t\mid s_t)$ and $\pi_\theta\ll\pi_{\theta_{\rm old}}$, the gradient at each prefix is
\begin{equation}
\nabla_\theta\KL\big(\pi_\theta(\cdot\mid s_t)\,\Vert\,\pi_T(\cdot\mid s_t)\big)
=-\E_{a_t\sim\pi_{\theta_{\rm old}}(\cdot\mid s_t)}\big[r_t(\theta)\,\Delta\log p_t\,\nabla_\theta\log\pi_\theta(a_t\mid s_t)\big].
\label{eq:klgrad}
\end{equation}
OPD is thus a policy-gradient update with the detached token advantage $A^{\rm OPD}_t=\sg[\Delta\log p_t]$, usually optimized with the PPO clipped surrogate \citep{ref38,ref4}. With $n$ trajectories $\{\tau_i\}_{i=1}^n$ per input and $N=\sum_i|\mathcal T(\tau_i)|$,
\begin{equation}
\mathcal L_{\rm clip}(\theta)=-\frac{1}{N}\sum_{i=1}^{n}\sum_{t\in\mathcal T(\tau_i)}\min\big\{r_{i,t}(\theta)A^{\rm OPD}_{i,t},\,\clip(r_{i,t}(\theta),1-\epsilon,1+\epsilon)A^{\rm OPD}_{i,t}\big\}.
\label{eq:opdloss}
\end{equation}
Vanilla OPD uses $n=1$; Multi-Rollout OPD uses $n>1$ independent trajectories. As in the local distillation update of \citet{ref75}, gradients do not flow through the sampled prefixes (derivations in Appendix~\ref{app:theory}).

\section{Method}
\label{sec:method}
\begin{figure}[!t]
\centering
\includegraphics[width=\linewidth]{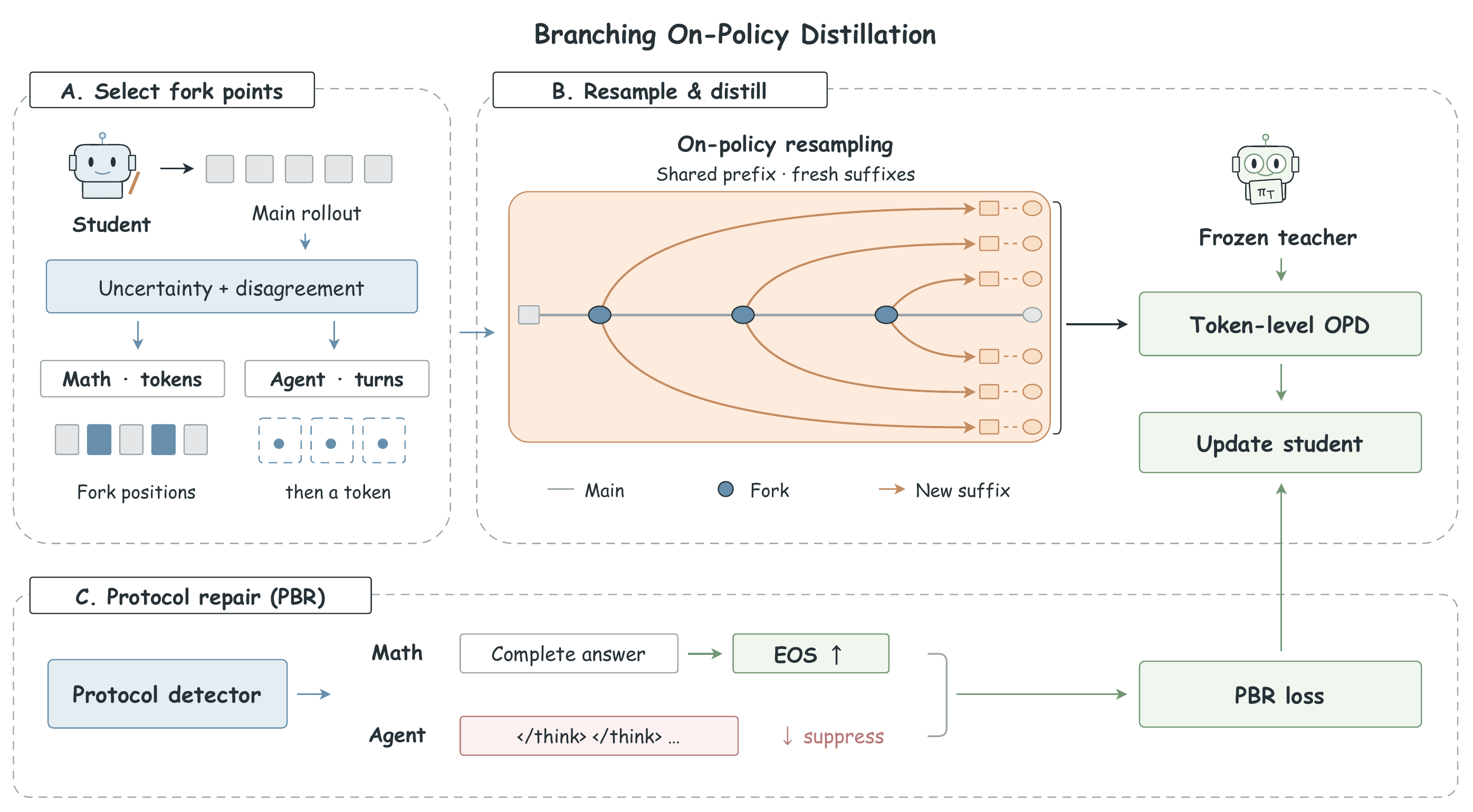}
\caption{Overview of \method{}. Fork points are selected on the main trajectory, suffixes are resampled from the student, and the teacher scores every generated token; shared prefixes are counted once. The bottom examples show concrete PBR constraints used in Section~\ref{sec:pbr_exp}.}
\label{fig:method}
\end{figure}

As Figure~\ref{fig:method} shows, \method{} selects fork points on a student trajectory, resamples continuations from them with the student policy, and trains on all generated tokens with the OPD loss and PBR.
\subsection{Selecting fork points}
For each input $x$, we sample a main trajectory $\tau_0\sim\pi_{\theta_{\rm old}}(\cdot\mid x)$ and select $B_{\rm eff}\leq B$ fork positions $t_1,\ldots,t_{B_{\rm eff}}\in\mathcal T(\tau_0)$ with prefixes $s_b=s_{t_b}$. From each prefix we independently sample $k$ new suffixes from $\pi_{\theta_{\rm old}}(\cdot\mid s_b)$, so each input yields at most $1+Bk$ trajectories. In agent tasks the environment is first restored to its state at $s_b$; in mathematics, generation simply continues from $s_b$. We keep every sampled suffix, including those whose first token repeats the main trajectory's, since they can still diverge later.

Candidate positions are scored with two signals. The uncertainty score $U_t$ is the student's entropy, computed from its top-20 token probabilities. The disagreement score $D_t$ aggregates $-\Delta\log p_t$ over a local window of 128 tokens, so it is large where the student prefers tokens that the teacher does not. Both scores are rank-normalized within the trajectory (denoted $R$) and mixed:
\begin{equation}
v_t=\alpha R(U_t)+(1-\alpha)R(D_t),\qquad
v_{\rm turn}(m)=\operatorname{mean}\big(\operatorname{TopFrac}_{\eta}\{v_t:t\in y_m\}\big),
\label{eq:score}
\end{equation}
where $\operatorname{TopFrac}_\eta$ keeps the top $\eta$ fraction of scores within turn $m$. In mathematics, we select up to $B$ forks among the highest-scoring eligible positions. In agent tasks, we select up to $B$ turns with the highest $v_{\rm turn}$, subject to a minimum gap between selected turns, and place each fork at a position drawn uniformly from the three highest-scoring eligible positions of its turn. By default, $\alpha=0.5$, $\eta=0.2$, $B=3$, and $k=2$; the scores are not used in the loss. Appendix~\ref{app:implementation} lists the eligibility rules and selection settings.

\subsection{Conditional resampling and coverage}
Let $\mathcal F_0$ collect the main trajectory and the selected forks, and let $\tau_{b,j}$ be the $j$-th suffix sampled from $s_b$. We fix $\pi_{\theta_{\rm old}}$ and $\theta$ before sampling the branches. For an integrable suffix statistic $f_\theta$, define
\begin{equation}
J_\theta(s_b)=\E_{\tau_{\geq t_b}\sim\pi_{\theta_{\rm old}}(\cdot\mid s_b)}f_\theta(\tau_{\geq t_b}),\qquad
\widehat J_{\rm RS}(s_b)=\frac1k\sum_{j=1}^k f_\theta(\tau_{b,j}).
\label{eq:rs}
\end{equation}
\begin{proposition}[Conditional unbiasedness]
\label{prop:rs}
If restored contexts have the correct continuation distribution and each branch satisfies $\tau_{b,j}\mid\mathcal F_0\sim\pi_{\theta_{\rm old}}(\cdot\mid s_b)$, then $\E[\widehat J_{\rm RS}(s_b)\mid\mathcal F_0]=J_\theta(s_b)$.
\end{proposition}
The result follows from linearity of conditional expectation (Appendix~\ref{app:theory}). If the branches are also conditionally independent with finite variance, $\Var(\widehat J_{\rm RS}\mid\mathcal F_0)=\Var(f_\theta\mid\mathcal F_0)/k$. Fork selection may depend on the main trajectory, but it is contained in $\mathcal F_0$ and therefore does not bias this estimate. Branching also widens coverage at the fork. Let $a_0$ be the main-trajectory token at $s_b$ and $a_{b,j}$ the first token of $\tau_{b,j}$. Then
\begin{align}
\Pr(a\text{ visited}\mid\mathcal F_0)&=1-[1-\pi_{\theta_{\rm old}}(a\mid s_b)]^k,\nonumber\\
\Pr(\exists j:a_{b,j}\ne a_0\mid\mathcal F_0)&=1-\pi_{\theta_{\rm old}}(a_0\mid s_b)^k.
\label{eq:coverage}
\end{align}
Both probabilities grow with $k$ but with diminishing increments, which motivates spreading the budget over several forks.

\subsection{Training objective}
The training batch contains the main trajectory and the $B_{\rm eff}k$ branches, indexed by $i$. The mask $M_{i,t}$ keeps all student-generated tokens of the main trajectory and, on each branch, only the tokens from the fork onward; observations and padding are masked out. We compute $\Delta\log p_{i,t}$ (Equation~\ref{eq:dlogp}) with the current parameters, apply a log-probability floor, and detach the result to form the advantage. With signal threshold $\delta$ and PPO clipping $\epsilon$ \citep{ref38}, Equation~\ref{eq:opdloss} becomes
\begin{align}
A_{i,t}&=\clip\!\left(\sg[\Delta\log p_{i,t}],-\delta,\delta\right),\quad
r_{i,t}(\theta)=\frac{\pi_\theta(a_{i,t}\mid s_{i,t})}{\pi_{\theta_{\rm old}}(a_{i,t}\mid s_{i,t})},\\
g_{i,t}&=-\min\{r_{i,t}A_{i,t},\clip(r_{i,t},1-\epsilon,1+\epsilon)A_{i,t}\},\\
L_{\rm tree}&=\frac{\sum_{i,t}M_{i,t}g_{i,t}}{\sum_{i,t}M_{i,t}}.
\label{eq:treeloss}
\end{align}
Because the loss averages over valid tokens, longer suffixes contribute more terms, and each shared prefix is counted once.

\subsection{Protocol-boundary repair}
\label{sec:pbr}
Some errors are defined by the task protocol rather than by the teacher: a mathematical response should end after its final answer, and an agent response should not begin with a stray closing-thought tag. OPD does not reliably remove them, because a teacher conditioned on a repetitive student prefix may continue the repetition itself (Appendix~\ref{app:pbr}). PBR supervises such events directly. Fixed detection rules scan each generated trajectory and emit constraints $(s,C,b)$: at prefix $s$, the next token should ($b=1$) or should not ($b=0$) lie in a nonempty token set $C\subsetneq\mathcal V$; detection is not differentiated. We define
\begin{align}
p_\theta(C\mid s)&=\sum_{v\in C}\pi_\theta(v\mid s),\nonumber\\
\ell_{\rm proto}(s,C,b)&=-b\log p_\theta(C\mid s)-(1-b)\log[1-p_\theta(C\mid s)],\\
L_{\rm train}&=L_{\rm tree}+\Psi\!\left(\lambda\operatorname{mean}_{(s,C,b)\in\mathcal C_{\rm batch}}\ell_{\rm proto}\right),
\label{eq:pbr}
\end{align}
where $\mathcal C_{\rm batch}$ is the set of constraints detected in the batch and $\lambda$ is the PBR weight; the auxiliary term is zero when $\mathcal C_{\rm batch}$ is empty. $\Psi$ is the identity, or $\Psi(u)=\min(u,c_{\rm PBR})$ when a loss cap is used. Positive constraints raise the probability of beneficial events and negative constraints lower that of harmful ones, as in unlikelihood training \citep{ref12}. PBR applies equally to main and branch trajectories. Appendix~\ref{app:algorithm} gives pseudocode.

\section{Experiments}
\label{sec:exp}
\subsection{Experimental setup}
We evaluate on mathematical reasoning and three agent benchmarks. For mathematics, we distill Qwen3-8B into Qwen3-4B-Base on DAPO-Math-17K \citep{ref15,ref16}. For the agent tasks, the student is Qwen3-4B; the teacher is Qwen3-30B-A3B on ALFWorld and a GRPO-trained checkpoint of the same model on WebShop and Search-QA. To test other student sizes and model families, we also distill Qwen3-4B into Qwen3-1.7B, and Qwen2.5-7B-RL into Qwen2.5-3B on ALFWorld. Benchmarks, prompts, and hyperparameters are described in Appendices~\ref{app:benchmarks}, \ref{app:prompts}, and~\ref{app:implementation}.

\paragraph{Evaluation.}
For mathematics, we report avg@16, the accuracy averaged over 16 sampled answers per problem, on MATH-500, AIME 2024, and AIME 2025. On ALFWorld \citep{ref13}, we report success rate (SR) on 140 seen (ID) and 134 unseen (OOD) tasks and their mean. On WebShop \citep{ref14}, we report SR on 128 held-out tasks, and on Search-QA, exact-match SR on the same mixed validation set for all models. Agents are evaluated with one trajectory per task; no branching is used at evaluation time.

\paragraph{Baselines.}
Our main baselines are Vanilla OPD, Multi-Rollout OPD with seven independent trajectories per input (matching the $1+Bk=7$ trajectories of \method{}), and Entropy-Weighted OPD, which reweights the token losses of a single trajectory. Table~\ref{tab:math} also includes offline KD, GRPO, EOPD \citep{ref58}, and SPOT \citep{ref79} (Appendix~\ref{app:math_baselines}); Table~\ref{tab:agents} includes KD, GRPO, and the curriculum method TCOD \citep{ref18}. Baselines trained with PBR are marked (+PBR).

\begin{table}[t]
\centering\small
\caption{Mathematics avg@16 (\%). Avg is the macro-average across the three evaluations. Baseline variants using protocol-boundary repair are marked (+PBR). }
\label{tab:math}
\fitTable{%
\begin{tabular}{lcccc}
\toprule
Method & MATH-500 & AIME24 & AIME25 & Avg \\
\midrule
Qwen3-8B teacher & 89.13 & 58.54 & 46.25 & 64.64 \\
Qwen3-4B-Base student & 48.91 & 8.75 & 7.50 & 21.72 \\
\midrule
KD & 74.73 & 12.50 & 12.08 & 33.10 \\
GRPO & 79.18 & 14.17 & 13.33 & 35.56 \\
Vanilla OPD & 64.33 & 7.08 & 6.04 & 25.82 \\
Multi-Rollout OPD ($n=7$) & 66.64 & 10.21 & 6.67 & 27.84 \\
Entropy-Weighted OPD & 67.94 & 12.29 & 8.96 & 29.73 \\
EOPD & 79.73 & 17.50 & 16.67 & 37.97 \\
SPOT & 79.88 & 17.50 & 17.92 & 38.43 \\
\midrule
Vanilla OPD (+PBR) & 79.13 & 19.58 & 18.13 & 38.94 \\
Multi-Rollout OPD ($n=7$, +PBR) & 78.94 & 19.79 & 18.54 & 39.09 \\
Entropy-Weighted OPD (+PBR) & 80.24 & 21.88 & 20.83 & 40.98 \\
\midrule
\rowcolor{branchblue}
\textbf{BranchOPD} & \textbf{81.24} & \textbf{26.04} & \textbf{25.83} & \textbf{44.37} \\
\bottomrule
\end{tabular}}
\end{table}

\begin{table}[t]
\centering\small
\caption{Performance on ALFWorld, Search-QA, and WebShop. We report success rate (SR, \%); ALFWorld reports ID SR and OOD SR. Avg. SR is the mean success rate over the three datasets, averaging ALFWorld ID and OOD first. $\dagger$: the WebShop and Search-QA teacher checkpoints are GRPO-trained.}
\label{tab:agents}
\fitTable{%
\begin{tabular}{lccccc}
\toprule
\multirow{2}{*}{Method} & \multicolumn{2}{c}{ALFWorld} & WebShop & Search-QA & \multirow{2}{*}{Avg. SR} \\
\cmidrule(lr){2-3}
 & ID SR & OOD SR & SR & SR &  \\
\midrule
Qwen3-30B-A3B & 53.57 & 50.00 & 31.25$^\dagger$ & 51.12$^\dagger$ & 44.72 \\
Qwen3-4B student & 35.71 & 35.82 & 2.34 & 28.71 & 22.27 \\
\midrule
KD & 38.57 & 38.06 & 18.75 & 43.50 & 33.52 \\
GRPO & 51.43 & 50.75 & 31.25 & 49.00 & 43.78 \\
Vanilla OPD & 42.14 & 43.28 & 25.78 & 48.24 & 38.91 \\
Multi-Rollout OPD ($n=7$) & 43.57 & 42.54 & 26.56 & 48.26 & 39.29 \\
Entropy-Weighted OPD & 46.43 & 46.27 & 27.34 & 48.56 & 40.75 \\
TCOD & 52.14 & 50.75 & 31.25 & 48.70 & 43.80 \\
\midrule
Vanilla OPD (+PBR) & 49.29 & 49.25 & 30.47 & 48.58 & 42.77 \\
Multi-Rollout OPD ($n=7$, +PBR) & 50.00 & 47.76 & 31.25 & 48.60 & 42.91 \\
Entropy-Weighted OPD (+PBR) & 52.86 & 51.49 & 32.03 & 48.90 & 44.37 \\
\midrule
\rowcolor{branchblue}
\textbf{BranchOPD} & \textbf{57.86} & \textbf{55.22} & \textbf{33.59} & \textbf{49.05} & \textbf{46.39} \\
\bottomrule
\end{tabular}}
\end{table}

\subsection{Main Results}
\method{} gives the best trained-student results in both tables, improving over Vanilla OPD by 18.55 points in mathematics Avg and 13.83 points in mean ALFWorld success (Tables~\ref{tab:math} and~\ref{tab:agents}). With PBR on every method, it reaches 44.37 mathematics Avg, 5.43 and 5.28 points above Vanilla OPD and Multi-Rollout OPD (n = 7), and above Entropy-Weighted OPD.

\paragraph{Mathematical reasoning.}
\method{} outperforms KD, GRPO, EOPD, and SPOT by 11.27, 8.81, 6.40, and 5.94 Avg points (Table~\ref{tab:math}). Gains are larger on AIME than on MATH-500: over SPOT, \method{} improves by 1.36 points on MATH-500 but by 8.54 and 7.91 points on AIME24 and AIME25. GRPO and SPOT use answer verification, whereas \method{} uses only teacher supervision.

\paragraph{Agent tasks.}
Averaged over the three benchmarks, \method{} reaches 46.39 SR, outperforming KD, GRPO, and TCOD by 12.87, 2.62, and 2.60 points, respectively (Table~\ref{tab:agents}). On ALFWorld, it raises mean success from 49.27\% for Vanilla OPD (+PBR) to 56.54\%, ahead of the other PBR baselines and 4.75 points above its teacher (51.79\%), without any task reward. On WebShop, it raises SR from 30.47\% for Vanilla OPD (+PBR) to 33.59\%, above the teacher (31.25\%). On Search-QA, the gain over Vanilla OPD (+PBR) is 0.47 points. Tables~\ref{tab:t15} and~\ref{tab:t16} report results for a smaller student and another model family.

\subsection{Ablations: Where and How Much to Branch}
\label{sec:ablation}
\begin{table}[t]
\centering\small
\caption{Ablations of fork selection and branch budget, all with PBR (ALFWorld SR and mathematics Avg, \%). \textit{Fork selection} rows share the eligible positions, budget, and resampling, and differ only in how forks are chosen. \emph{Random}: uniform over eligible positions, without scoring (agent tasks keep the minimum turn gap). \emph{Entropy only} / \emph{Disagreement only}: $v_t$ in Equation~\ref{eq:score} with $\alpha=1$ / $\alpha=0$. \emph{Turn-initial} (agent tasks only): turns chosen by the same $v_{\rm turn}$, but each fork is placed at the turn's first generated token. \textit{Branch budget} rows vary the number of forks $B$ and continuations per fork $k$ ($1+Bk$ trajectories per input).}
\label{tab:main_ablation}
\fitTable{%
\begin{tabular}{lccc}
\toprule
Setting & \multicolumn{2}{c}{ALFWorld} & Math Avg \\
\cmidrule(lr){2-3}
 & ID SR & OOD SR & \\
\midrule
\rowcolor{pbrgray}
\multicolumn{4}{l}{\textit{Fork selection ($B=3,k=2$)}} \\
Random & 50.71 & 46.27 & 40.11 \\
Entropy only ($\alpha=1$) & 55.71 & 52.99 & 43.24 \\
Disagreement only ($\alpha=0$) & 55.00 & 52.24 & 42.97 \\
Turn-initial & 55.71 & 53.73 & -- \\
\rowcolor{branchblue}
\textbf{Entropy + disagreement ($\alpha=0.5$)} & \textbf{57.86} & \textbf{55.22} & \textbf{44.37} \\
\addlinespace[3pt]
\rowcolor{pbrgray}
\multicolumn{4}{l}{\textit{Branch budget (entropy + disagreement)}} \\
$B=3, k=1$ & 54.29 & 50.75 & 42.18 \\
$B=3, k=3$ & 57.14 & 54.48 & 43.91 \\
$B=2, k=2$ & 57.14 & 54.48 & 43.94 \\
\midrule
\rowcolor{branchblue}
\textbf{$B=3, k=2$ (default)} & \textbf{57.86} & \textbf{55.22} & \textbf{44.37} \\
\bottomrule
\end{tabular}}
\end{table}

\paragraph{Where to branch.}
The caption of Table~\ref{tab:main_ablation} defines each selection rule. Mixing entropy and disagreement works best. Compared with random forks, it improves ALFWorld ID/OOD SR by 7.14/8.96 points and raises mathematics Avg from 40.11 to 44.37. Either signal alone recovers most, but not all, of this gain. Turn-initial branching trails the default by 2.14/1.49 points on ALFWorld, so where the fork falls within a turn also matters. Table~\ref{tab:t8} reports further selection and filtering controls.

\paragraph{How much to branch.}
With $B=3$, increasing the continuations per fork from $k=1$ to $k=2$ improves all three metrics, while $k=3$ brings no further gain. With $k=2$, going from two to three forks adds 0.43 mathematics points and 0.73 ALFWorld points, so it pays to cover several decisions with a few continuations each (Table~\ref{tab:t7}).

\subsection{Behavior Coverage and Sampling Efficiency}
\label{sec:coverage}
Behavior coverage is the range of alternative behaviors the student explores from the same decision state. We measure both the probability mass left to alternatives at a fork, $1-\pi_{\theta_{\rm old}}(a_0\mid s)$ with $a_0$ the main-trajectory token, and the behavioral differences that branches actually produce. At the selected forks (step 100), this mass is about 0.63 on ALFWorld and 0.69 on WebShop, and branches do produce different behavior: 74.2\% and 47.7\% take a different action sequence from the main trajectory, and 55.5\% and 37.5\% contain an action absent from it (Table~\ref{tab:t5}). Branching thus turns mass that a single rollout leaves unrealized into additional histories with teacher feedback. Branches can also recover from failures: when the main trajectory fails, at least one branch succeeds in 5.0\% (WebShop) to 42.7\% (mathematics) of cases (Table~\ref{tab:t6}). Appendix~\ref{app:diagnostics} tracks these statistics over training.

\begin{figure}[t]
\centering\includegraphics[width=\linewidth]{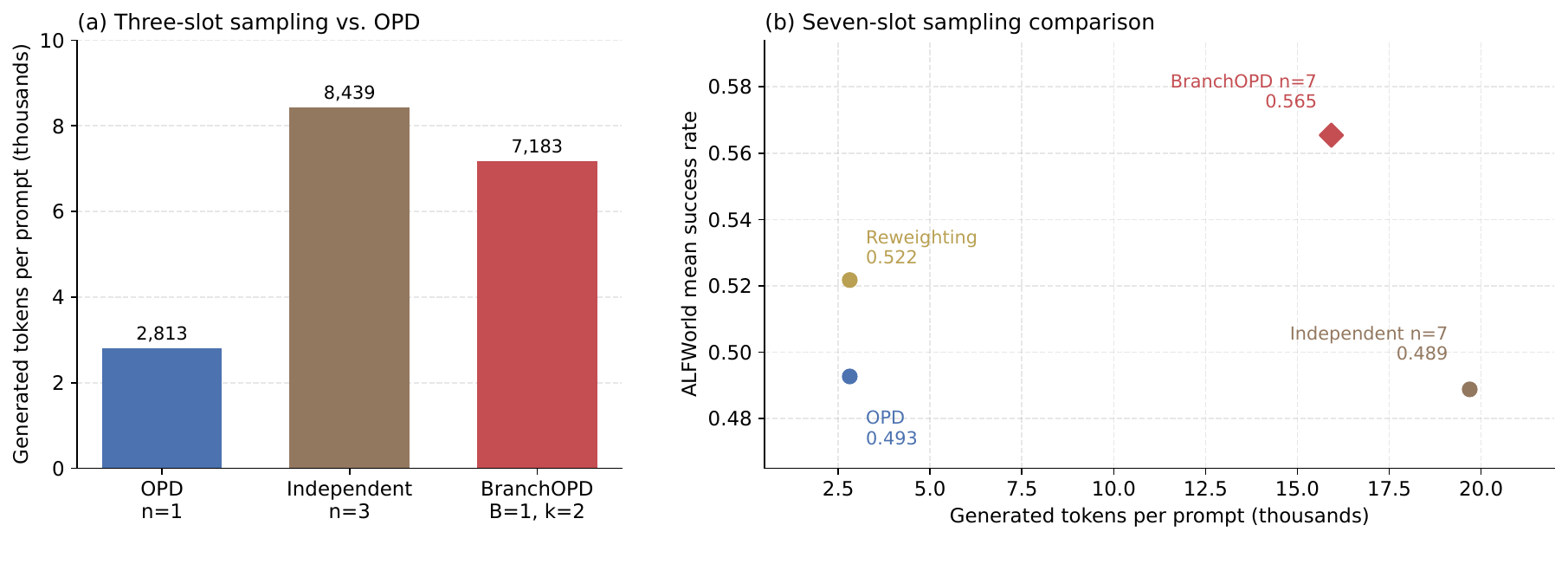}
\caption{ALFWorld generation cost and success with PBR, for three (left) and seven (right) trajectory slots. OPD, Independent, and Reweighting denote Vanilla OPD, Multi-Rollout OPD, and Entropy-Weighted OPD.}
\label{fig:budget}
\end{figure}

While achieving broader behavior coverage, BranchOPD requires fewer student-generated tokens by reusing the prefix shared with the main trajectory. On ALFWorld, one trajectory requires 2,813 student tokens on average; seven independent trajectories would therefore require 19,691 tokens, whereas our seven-trajectory branching setting uses only 15,924 tokens, saving 19.1\%. On WebShop, the corresponding cost is 4,824 tokens versus 7,014 for seven independent trajectories, saving 31.2\% (Figure~\ref{fig:budget}; Table~\ref{tab:t10}). These counts exclude teacher scoring, backpropagation, and environment replay.

\subsection{Protocol-Boundary Repair}
\label{sec:pbr_exp}
\begin{figure}[t]
\centering\includegraphics[width=\linewidth]{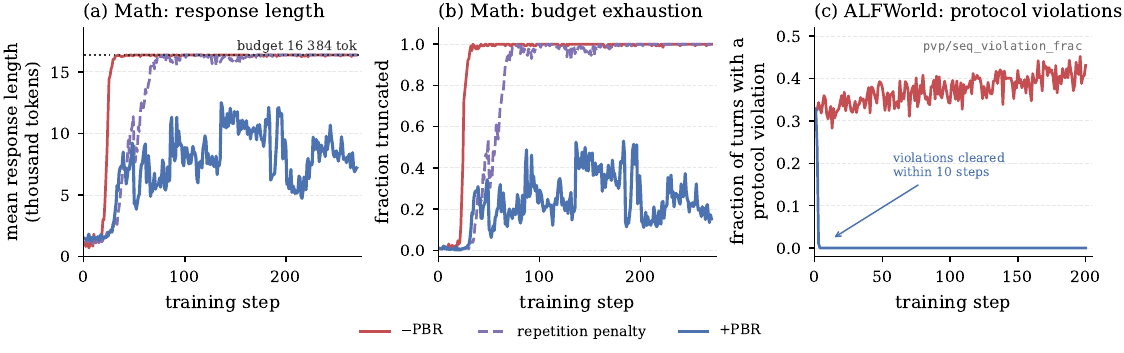}
\caption{Training dynamics with PBR: mathematics response length and truncation rate (fraction of responses stopped at the length limit), and ALFWorld turn-level protocol violation (fraction of turns with a violation). Repetition penalties delay budget exhaustion but leave generation near the cap; PBR reduces both truncation and violations.}
\label{fig:pbr}
\end{figure}

PBR explicitly controls protocol events that should either be encouraged or suppressed. For example, it encourages termination after a complete mathematical answer by setting (C=$\mathrm{EOS}$) and (b=1), while suppressing extra leading closing-thought tags in agent responses with (b=0). On mathematics, PBR alone lifts Vanilla OPD from 25.82 to 38.94 Avg (Table~\ref{tab:math}) and leaves more room below the length limit (Figure~\ref{fig:pbr}). On ALFWorld, it improves ID/OOD SR by 7.14/5.97 points and removes the extra leading tags entirely. A teacher that does not repeat in its own rollouts can still reinforce repetition when conditioned on a repetitive student prefix, so the repair must act on student prefixes (Appendix~\ref{app:pbr}).

\paragraph{Why targeted repair?}
We also compare PBR with generic remedies that broadly discourage repetition or truncate problematic samples, which are consistently less effective than PBR (Table~\ref{tab:t11}). Repetition penalties achieve only 28.70--30.40 Avg, filtering overlong samples reaches 33.20, and early stopping reaches 22.05, compared with 38.94 for PBR. Moreover, repetition penalties mainly slow the increase in response length rather than preventing it from eventually approaching the generation limit (Figure~\ref{fig:pbr}).

\section{Conclusion}
We presented \method{}, which resamples student continuations at selected decision points to broaden on-policy distillation, together with protocol-boundary repair. It outperforms Vanilla OPD and Multi-Rollout OPD on every evaluated benchmark while generating fewer student tokens than independent rollouts. What a student learns from OPD depends not only on the teacher signal but also on which of its own alternative histories it gets to see.

\label{main:end}
\clearpage
\subsection*{AI Use Statement}
We used large language models to assist with literature retrieval and synthesis, paper drafting and translation, refinement of theoretical formulations and proofs, and interpretation of results. The authors are responsible for the accuracy, originality, and final content of the submission, including AI-assisted text, claims, and artifacts.

\subsection*{Ethics Statement}
Our research uses public benchmarks and controlled task environments, and does not involve human-subject experiments or collection of sensitive personal data.

\bibliography{references}
\bibliographystyle{iclr2027_conference}
\clearpage
\appendix
\section{Theoretical Details}
\label{app:theory}
Throughout, the rollout policy $\pi_{\theta_{\rm old}}$ is fixed within each sampling group, $s$ denotes a prefix $s_t$, and
\begin{equation}
\Delta\log p(s,a)=\log\pi_T(a\mid s)-\log\pi_\theta(a\mid s)
\end{equation}
is the distillation signal of Equation~\ref{eq:dlogp} at token $a$. We assume $\pi_T(a\mid s)>0$ whenever $\pi_\theta(a\mid s)>0$, and $\pi_\theta\ll\pi_{\theta_{\rm old}}$ at every supervised prefix.

\subsection{OPD gradient and token surrogate}
At a fixed prefix $s$, the reverse KL is $\KL(\pi_\theta(\cdot\mid s)\Vert\pi_T(\cdot\mid s))=\sum_a\pi_\theta(a\mid s)[-\Delta\log p(s,a)]$. Using $\nabla_\theta\pi_\theta=\pi_\theta\nabla_\theta\log\pi_\theta$ and $\nabla_\theta[-\Delta\log p(s,a)]=\nabla_\theta\log\pi_\theta(a\mid s)$,
\begin{align}
\nabla_\theta\KL(\pi_\theta(\cdot\mid s)\Vert\pi_T(\cdot\mid s))
&=\sum_a\pi_\theta(a\mid s)\big(1-\Delta\log p(s,a)\big)\nabla_\theta\log\pi_\theta(a\mid s)\\
&=-\sum_a\pi_\theta(a\mid s)\,\Delta\log p(s,a)\,\nabla_\theta\log\pi_\theta(a\mid s),
\end{align}
since $\sum_a\pi_\theta\nabla_\theta\log\pi_\theta=\nabla_\theta\sum_a\pi_\theta=0$. Under $\pi_\theta\ll\pi_{\theta_{\rm old}}$, rewriting the sum as an expectation over $a\sim\pi_{\theta_{\rm old}}(\cdot\mid s)$ with weight $r(\theta)=\pi_\theta(a\mid s)/\pi_{\theta_{\rm old}}(a\mid s)$ gives Equation~\ref{eq:klgrad}. With $A^{\rm OPD}=\sg[\Delta\log p(s,a)]$, the unclipped surrogate $-r(\theta)A^{\rm OPD}$ has gradient $-A^{\rm OPD}r(\theta)\nabla_\theta\log\pi_\theta(a\mid s)$, whose expectation under $\pi_{\theta_{\rm old}}$ equals the KL gradient when the detached signal is evaluated at the differentiation point. PPO and signal clipping then define the training surrogates in Equations~\ref{eq:opdloss} and~\ref{eq:treeloss}.

Summing over $t\in\mathcal T(\tau)$ and averaging over $\tau\sim\pi_{\theta_{\rm old}}$ yields the gradient of Equation~\ref{eq:opd} with the prefix distribution held fixed; the update does not differentiate through the rollout distribution \citep{ref75}. Equation~\ref{eq:opd} normalizes within each trajectory, whereas the token-mean surrogate normalizes over the batch and thus weights trajectories by length; the two coincide for a single trajectory.

\subsection{Conditional resampling}
Let $\mathcal F_0$ contain the main trajectory and the selected fork prefixes. At a fork prefix $s_b=s_{t_b}$, write a fresh continuation as $\tau_{\geq t_b}=(a,\tau_{>t_b})$, where $a$ is its first token and $\tau_{>t_b}$ includes subsequent tokens and environment transitions. For an integrable statistic $f_\theta$ fixed before drawing the continuations,
\begin{align}
\pi_{\theta_{\rm old}}(\tau_{\geq t_b}\mid s_b)&=\pi_{\theta_{\rm old}}(a\mid s_b)\,\pi_{\theta_{\rm old}}(\tau_{>t_b}\mid s_b,a),\\
\mu_\theta(s_b,a)&=\E_{\tau_{>t_b}\sim\pi_{\theta_{\rm old}}(\cdot\mid s_b,a)}f_\theta(a,\tau_{>t_b}),\\
J_\theta(s_b)&=\sum_a\pi_{\theta_{\rm old}}(a\mid s_b)\,\mu_\theta(s_b,a).
\end{align}
Restoring the fork state and drawing each continuation from $\pi_{\theta_{\rm old}}(\cdot\mid s_b)$ gives
\begin{equation}
\E[\widehat J_{\rm RS}(s_b)\mid\mathcal F_0]
=\frac1k\sum_{j=1}^k\E[f_\theta(\tau_{b,j})\mid\mathcal F_0]=J_\theta(s_b).
\end{equation}
This proves Proposition~\ref{prop:rs} using only the conditional marginal law of each fresh branch. For conditionally independent branches with finite variance,
\begin{equation}
\Var(\widehat J_{\rm RS}(s_b)\mid\mathcal F_0)=\frac{\Var(f_\theta(\tau_{\geq t_b})\mid\mathcal F_0)}{k}.
\end{equation}
Fork selection depends only on the main trajectory and is therefore contained in $\mathcal F_0$; the expectation above is taken over the fresh branch draws alone.

\subsection{Coverage and diminishing returns}
A token $a$ is absent from $k$ independent branch starts at $s_b$ with probability $(1-\pi_{\theta_{\rm old}}(a\mid s_b))^k$. Therefore, for the number $N_k$ of distinct first tokens among those branches,
\begin{align}
\E[N_k\mid\mathcal F_0]&=\sum_{a\in\mathcal V}\left[1-(1-\pi_{\theta_{\rm old}}(a\mid s_b))^k\right],\\
\E[N_{k+1}-N_k\mid\mathcal F_0]&=\sum_{a\in\mathcal V}\pi_{\theta_{\rm old}}(a\mid s_b)(1-\pi_{\theta_{\rm old}}(a\mid s_b))^k.
\end{align}
The marginal increase is nonnegative and decreases with $k$. For the realized main token $a_0$, at least one branch starts differently with probability $1-\pi_{\theta_{\rm old}}(a_0\mid s_b)^k$. Branches that share the first token can still diverge later, so we measure both first-token alternatives and later behavioral differences (Appendix~\ref{app:diagnostics}).

\subsection{Token aggregation and protocol repair}
Proposition~\ref{prop:rs} concerns the average of a fixed suffix statistic, whereas the training loss averages over valid tokens in the main trajectory and all branches. A suffix with $L$ generated tokens contributes $L$ loss terms, and shared prefix tokens enter once. Fork selection thus determines which prefixes are supervised, continuation length determines how much each contributes, and the mask $M_{i,t}$ sets the normalization.

For PBR, a positive event contributes $-\log p_\theta(C\mid s)$ and a negative event $-\log(1-p_\theta(C\mid s))$; their gradients increase and decrease the event probability, respectively. Detection is held fixed during differentiation. With the cap $\Psi(u)=\min(u,c_{\rm PBR})$, the auxiliary gradient is unchanged below the cap and zero above it; a batch without constraints contributes zero.

\subsection{Weighted forced-token comparator}
The forced-token controls in Table~\ref{tab:t9} use a natural token plus $k$ alternatives. At a fixed prefix $s$, write $p_{\rm old}(a)=\pi_{\theta_{\rm old}}(a\mid s)$, sample $A_0\sim p_{\rm old}$, and let
\begin{equation}
S(A_0)=\{A_0\}\cup\operatorname{TopK}_{a\ne A_0}p_{\rm old}(a).
\end{equation}
With deterministic tie-breaking and at least $k+1$ tokens in the support, order tokens as $a_1,a_2,\ldots$ by decreasing $p_{\rm old}$ and write $Q_{k+1}=\sum_{i=1}^{k+1}p_{\rm old}(a_i)$. Their inclusion probabilities are
\begin{equation}
\rho(a_i)=\begin{cases}1,&i\leq k,\\Q_{k+1},&i=k+1,\\p_{\rm old}(a_i),&i>k+1.\end{cases}
\end{equation}
The first $k$ tokens are always included; token $k+1$ enters when $A_0$ is among the first $k+1$ tokens; all others enter through the natural draw. After forcing token $a$, the continuation $\tau^{(a)}_{>t}\sim\pi_{\theta_{\rm old}}(\cdot\mid s,a)$ has conditional mean $\mu_\theta(s,a)$. Horvitz--Thompson weights \citep{ref77} then give
\begin{align}
\widehat J_{\rm HT}(s)&=\sum_{a\in S(A_0)}\frac{p_{\rm old}(a)}{\rho(a)}f_\theta(a,\tau^{(a)}_{>t}),\\
\E[\widehat J_{\rm HT}(s)\mid s]&=\sum_a\frac{p_{\rm old}(a)}{\rho(a)}\rho(a)\mu_\theta(s,a)=J_\theta(s).
\end{align}
Every token with positive probability has positive inclusion probability. The scaled weights $w_{\rm scaled}(a)=(k+1)\,p_{\rm old}(a)/\rho(a)$ satisfy $\widehat J_{\rm HT}=(k+1)^{-1}\sum_{a\in S}w_{\rm scaled}(a)f_\theta(a,\tau^{(a)}_{>t})$ and $\E\sum_{a\in S}w_{\rm scaled}(a)=k+1$.

Table~\ref{tab:t9} uses scaled HT weights in a pooled token loss, with the main-slot weight averaged across its forks. RB instead self-normalizes the weights within each realized group; because its denominator is random, it is a different estimator. The HT identity above holds when the natural draw and the continuations follow the fixed-prefix sampling distribution.
\FloatBarrier

\clearpage
\section{Additional Results and Controlled Ablations}
\label{app:ablation}
This section reports results for other students, sampling rules, generation budgets, and protocol repair, using the metrics of Section~\ref{sec:exp}.
\Needspace{0.30\textheight}
\subsection{Student size and model family}
\begin{table}[H]
\centering\small
\caption{Qwen3-4B to Qwen3-1.7B distillation. Teachers are task-specific GRPO checkpoints. Values are task SR or mathematics mean avg@16 (\%); (+PBR) identifies repaired baselines.}
\label{tab:t15}
\fitTable{%
\begin{tabular}{lccccc}
\toprule
\multirow{2}{*}{Method} & \multicolumn{2}{c}{ALFWorld} & WebShop & Search-QA & \multirow{2}{*}{Math Avg} \\
\cmidrule(lr){2-3}
 & ID SR & OOD SR & SR & SR &  \\
\midrule
Qwen3-4B GRPO teacher & 46.43 & 41.04 & 27.34 & 47.20 & 61.60 \\
Qwen3-1.7B student & 7.86 & 6.72 & 3.91 & 26.40 & 11.08 \\
\midrule
Vanilla OPD (+PBR) & 44.29 & 42.54 & 26.56 & 45.10 & 30.18 \\
Multi-Rollout OPD ($n=7$, +PBR) & 45.00 & 40.30 & 26.56 & 45.40 & 30.05 \\
Entropy-Weighted OPD (+PBR) & 47.14 & 45.52 & 27.34 & 45.90 & 31.60 \\
\midrule
\rowcolor{branchblue}
\textbf{BranchOPD} & \textbf{51.43} & \textbf{48.51} & \textbf{29.69} & \textbf{46.80} & \textbf{34.86} \\
\bottomrule
\end{tabular}}
\end{table}

\begin{table}[H]
\centering\small
\caption{Qwen2.5-7B-RL to Qwen2.5-3B on ALFWorld. SR (\%) uses 140 ID and 134 OOD tasks; trained baselines use PBR.}
\label{tab:t16}
\fitTable{%
\begin{tabular}{lcc}
\toprule
\multirow{2}{*}{Method} & \multicolumn{2}{c}{ALFWorld} \\
\cmidrule(lr){2-3}
 & ID SR & OOD SR \\
\midrule
Qwen2.5-7B-RL teacher & 64.29 & 55.22 \\
Qwen2.5-3B zero-shot & 7.14 & 2.99 \\
\midrule
Vanilla OPD (+PBR) & 47.14 & 41.79 \\
Multi-Rollout OPD ($n=7$, +PBR) & 47.86 & 40.30 \\
Entropy-Weighted OPD (+PBR) & 50.71 & 44.03 \\
\midrule
\rowcolor{branchblue}
\textbf{BranchOPD} & \textbf{55.71} & \textbf{50.75} \\
\bottomrule
\end{tabular}}
\end{table}

With Qwen3-1.7B as the student, \method{} improves over Vanilla OPD (+PBR) by 6.56 points on ALFWorld and 4.68 points on mathematics, and exceeds the teacher on ALFWorld and WebShop without task rewards (Table~\ref{tab:t15}). With Qwen2.5-3B, the ALFWorld gain is 8.76 points (Table~\ref{tab:t16}).
\FloatBarrier

\Needspace{0.30\textheight}
\subsection{Selection and branching budgets}
\begin{table}[H]
\centering\small
\caption{Selection and eligibility ablations with $B=3,k=2$ and PBR. ALFWorld SR and mathematics mean avg@16 are percentages. $\alpha$ weights entropy in the mixed selector; -- denotes an unevaluated setting.}
\label{tab:t8}
\fitTable{%
\begin{tabular}{lccc}
\toprule
\multirow{2}{*}{Setting} & \multicolumn{2}{c}{ALFWorld} & \multirow{2}{*}{Math Avg} \\
\cmidrule(lr){2-3}
 & ID SR & OOD SR &  \\
\midrule
Random sites & 50.71 & 46.27 & 40.11 \\
Student entropy only & 55.71 & 52.99 & 43.24 \\
Disagreement only & 55.00 & 52.24 & 42.97 \\
Entropy + disagreement ($\alpha=0.3$) & 57.14 & 54.48 & 43.88 \\
\textbf{Entropy + disagreement ($\alpha=0.5$)} & \textbf{57.86} & \textbf{55.22} & \textbf{44.37} \\
Entropy + disagreement ($\alpha=0.7$) & 56.43 & 53.73 & 43.65 \\
No mathematics-aware eligibility & -- & -- & 42.03 \\
Turn-initial & 55.71 & 53.73 & -- \\
Failed mains only & 55.00 & 50.75 & 36.82 \\
\bottomrule
\end{tabular}}
\end{table}

\begin{table}[H]
\centering\small
\caption{Branch budgets with entropy + disagreement selection and PBR. Performance is in percent; slots equal $1+Bk$. Generation is relative to Vanilla OPD on ALFWorld; $\dagger$ marks interpolated or extrapolated costs.}
\label{tab:t7}
\fitTable{%
\begin{tabular}{lccccc}
\toprule
\multirow{2}{*}{Setting} & \multicolumn{2}{c}{ALFWorld} & \multirow{2}{*}{Math Avg} & \multirow{2}{*}{Slots} & \multirow{2}{*}{Generation} \\
\cmidrule(lr){2-3}
 & ID SR & OOD SR &  &  &  \\
\midrule
B=3,k=1 & 54.29 & 50.75 & 42.18 & 4 & 3.33$\times$ $\dagger$ \\
\textbf{B=3,k=2} & \textbf{57.86} & \textbf{55.22} & \textbf{44.37} & 7 & \textbf{5.66$\times$} \\
B=3,k=3 & 57.14 & 54.48 & 43.91 & 10 & 7.99$\times$ $\dagger$ \\
B=1,k=2 & 54.29 & 51.49 & 42.55 & 3 & 2.55$\times$ \\
B=2,k=2 & 57.14 & 54.48 & 43.94 & 5 & 4.11$\times$ $\dagger$ \\
\bottomrule
\end{tabular}}
\end{table}

Entropy + disagreement is the best of the tested selection rules (Table~\ref{tab:t8}). Disabling the mathematics-aware eligibility filter lowers mathematics Avg to 42.03, and branching only from failed main trajectories lowers it to 36.82, so branches from successful trajectories are useful as well.

The two budgets play different roles (Table~\ref{tab:t7}). Going from $k=1$ to $k=2$ helps, and $k=3$ adds nothing. At $k=2$, increasing $B$ from one to three raises mathematics Avg from 42.55 to 44.37. The five-trajectory setting ($B=2,k=2$) keeps most of the gain at a lower generation cost.
\FloatBarrier

\Needspace{0.30\textheight}
\subsection{Sampling and perturbation controls}
\label{app:expansion}
\begin{table}[H]
\centering\small
\caption{Expansion rules with PBR. Performance is in percent. HT uses scaled inclusion weights; RB normalizes weights within each sampled group (Appendix~\ref{app:theory}).}
\label{tab:t9}
\fitTable{%
\begin{tabular}{lcccccc}
\toprule
\multirow{2}{*}{Expansion} & \multirow{2}{*}{Weights} & \multicolumn{2}{c}{ALFWorld} & \multirow{2}{*}{Math Avg} & WebShop & Search-QA \\
\cmidrule(lr){3-4}
 &  & ID SR & OOD SR &  & SR & SR \\
\midrule
forced-topk & Uniform & 52.86 & 48.51 & 40.90 & 31.25 & 48.76 \\
forced-topk & RB & 57.14 & 51.49 & 41.77 & 31.25 & \textbf{49.22} \\
forced-topk & HT (scaled) & 55.71 & 50.75 & 42.11 & 32.03 & 48.96 \\
\midrule
\rowcolor{branchblue}
\textbf{resample} & Uniform & \textbf{57.86} & \textbf{55.22} & \textbf{44.37} & \textbf{33.59} & 49.05 \\
\bottomrule
\end{tabular}}
\end{table}

\begin{table}[H]
\centering\small
\caption{ALFWorld continuation diagnostics. The first five metrics use step 100; low-likelihood rates summarize 64-token windows. Rates are percentages and new actions is a count.}
\label{tab:t12}
\fitTable{%
\begin{tabular}{lcc}
\toprule
Metric & Forced + HT & Resample \\
\midrule
Main-token probability & 37.19\% & 36.67\% \\
Alternative mass & 62.81\% & 63.33\% \\
Action disagreement & 75.8\% & 74.2\% \\
New-action coverage (NSR) & 58.6\% & 55.5\% \\
Mean new actions & 2.59 & 2.44 \\
Low-likelihood suffixes & 19.2\% & 5.8\% \\
\bottomrule
\end{tabular}}
\end{table}

Among the expansion rules, resampling gives the highest ALFWorld, mathematics, and WebShop scores, and RB the highest Search-QA SR (Table~\ref{tab:t9}). Both rules change actions at similar rates, but resampling produces far fewer low-likelihood suffixes, 5.8\% versus 19.2\% (Table~\ref{tab:t12}). A suffix counts as low-likelihood when its mean student log probability over 64 tokens falls below the fifth percentile of on-policy windows of the same length. Useful coverage can therefore be obtained from continuations that the student is likely to generate.
\begin{table}[H]
\centering\small
\caption{Noise-token branching on ALFWorld with PBR. ID/OOD SR and their mean are percentages. Each branch starts with the forced token \texttt{!}.}
\label{tab:t13}
\fitTable{%
\begin{tabular}{lccc}
\toprule
\multirow{2}{*}{Setting} & \multicolumn{2}{c}{ALFWorld} & \multirow{2}{*}{ALF Avg} \\
\cmidrule(lr){2-3}
 & ID SR & OOD SR &  \\
\midrule
Vanilla OPD (+PBR) & 49.29 & 49.25 & 49.27 \\
Two-stage + noise token & 50.71 & 50.00 & 50.36 \\
Turn-initial + noise & 50.00 & 50.00 & 50.00 \\
\bottomrule
\end{tabular}}
\end{table}

Forcing a noise token (\texttt{!}) at the start of each branch yields small ALFWorld gains of 1.09 and 0.73 points over Vanilla OPD (+PBR), depending on where the forks are placed (Table~\ref{tab:t13}). Perturbed continuations still add some supervision, but the much larger gains of informative branching show that fork selection matters.
\FloatBarrier

\Needspace{0.30\textheight}
\subsection{Generation and environment costs}
\begin{table}[H]
\centering\small
\caption{Generation costs and performance with PBR. Scores are mathematics Avg or task SR (\%); ALFWorld averages ID/OOD. Costs are relative to Vanilla OPD, and savings use the $7\times$ independent-rollout reference.}
\label{tab:t10}
\fitTable{%
\begin{tabular}{lcccccc}
\toprule
Task & \shortstack{Vanilla\\OPD} & \shortstack{Multi-Rollout\\OPD ($n=7$)} & BranchOPD & Generation & Saving vs. $7\times$ & Cost basis \\
\midrule
Mathematics & 38.94 & 39.09 & 44.37 & 5.68$\times$ & 18.9\% & Position estimate \\
ALFWorld & 49.27 & 48.88 & 56.54 & 5.66$\times$ & 19.1\% & Measured tokens \\
WebShop & 30.47 & 31.25 & 33.59 & 4.81$\times$ & 31.2\% & Measured tokens \\
Search-QA & 48.58 & 48.60 & 49.05 & 4.24$\times$ & 39.4\% & Position estimate \\
\bottomrule
\end{tabular}}
\end{table}

\begin{table}[H]
\centering\small
\caption{Three-slot ALFWorld sampling with PBR. SR is the ID/OOD mean (\%); generation and environment steps are relative to Vanilla OPD.}
\label{tab:t14}
\fitTable{%
\begin{tabular}{lcccc}
\toprule
Setting & Slots & Generation & Env. steps & ALF Avg \\
\midrule
Vanilla OPD (+PBR) & 1 & 1.00$\times$ & 1.00$\times$ & 49.27 \\
Resample ($B=1,k=2$) & 3 & 2.55$\times$ & 3.02$\times$ & 52.89 \\
Forced, turn-initial + RB ($B=1,k=2$) & 3 & 2.49$\times$ & 3.07$\times$ & 54.48 \\
\bottomrule
\end{tabular}}
\end{table}

Relative to seven independent rollouts, \method{} saves 19.1\% and 31.2\% of student generation on ALFWorld and WebShop while improving SR; estimates from fork positions give savings of 18.9\% on mathematics and 39.4\% on Search-QA (Table~\ref{tab:t10}). With two added continuations, ALFWorld uses $2.55\times$ the generation and $3.02\times$ the environment steps of Vanilla OPD, because state replay adds environment cost on top of suffix generation (Table~\ref{tab:t14}).
\FloatBarrier

\Needspace{0.30\textheight}
\subsection{Protocol repair and repetition}
\label{app:pbr}
\begin{table}[H]
\centering\small
\caption{Repetition treatments for Vanilla OPD on mathematics. Values are mean avg@16 (\%).}
\label{tab:t11}
\fitTable{%
\begin{tabular}{lc}
\toprule
Recipe & Math Avg \\
\midrule
No PBR & 25.82 \\
Repetition penalty & 28.70 \\
n-gram / line penalty in advantage & 30.40 \\
Overlong-sample filtering & 33.20 \\
Early stopping during training & 22.05 \\
\midrule
\rowcolor{branchblue}
\textbf{Full PBR} & \textbf{38.94} \\
\bottomrule
\end{tabular}}
\end{table}

Without PBR, mathematics responses reach the length cap around step 30. Repetition penalties delay this but do not prevent persistent budget exhaustion (Figure~\ref{fig:pbr}). PBR reaches 38.94 mathematics Avg, compared with 33.20 for filtering and 28.70--30.40 for penalties; after step 50, its median response length is about 8,600 tokens and its median truncation rate is 0.28.

A teacher whose own rollouts are clean can still reinforce repetition when queried on a repetitive student prefix; PBR targets these prefix-conditioned errors. On ALFWorld, teacher responses contain on average 1.64/1.50 excess leading \texttt{</think>} markers on ID/OOD. Vanilla OPD raises this to 1.94/1.91, and PBR reduces it to zero. A single response can contain several such markers.
\begin{figure}[H]
\centering\includegraphics[width=\linewidth]{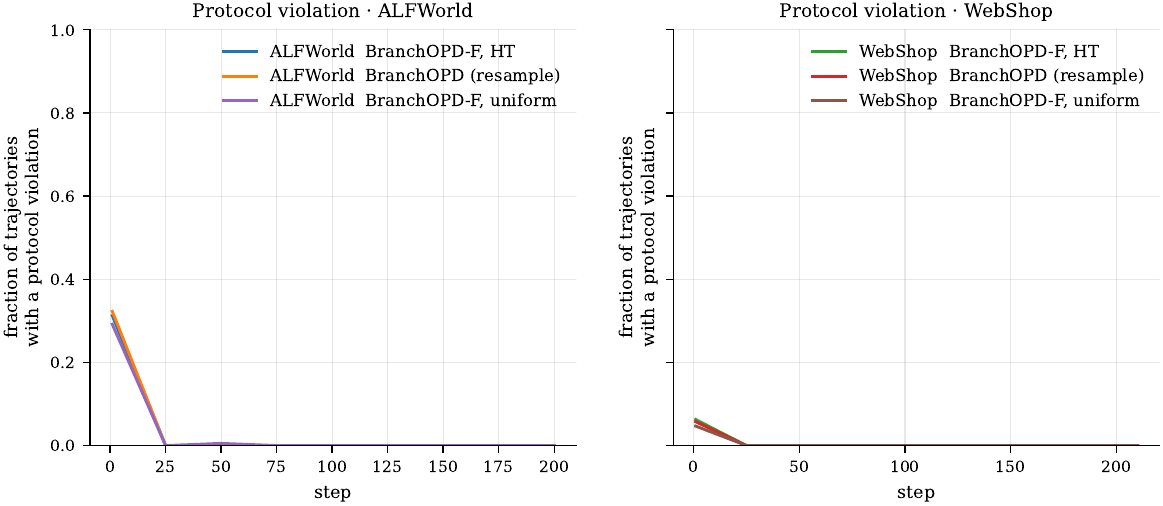}
\caption{Trajectory-level protocol violation (PVP): the fraction of the 192 main and branch trajectories that contain a violation. All methods shown use PBR; BranchOPD-F denotes forced-token expansion (Table~\ref{tab:t9}).}
\label{fig:pvp}
\end{figure}
Trajectory-level protocol violation drops to zero by step 25 and stays near zero (Figure~\ref{fig:pvp}). Together with the behavior coverage in Appendix~\ref{app:diagnostics}, this indicates that protocol repair does not suppress exploration.
\FloatBarrier

\Needspace{0.35\textheight}
\section{Behavior Coverage and Recovery}
\label{app:diagnostics}
These diagnostics follow the chain from selecting uncertain decisions, to generating alternative behavior, to recovering successful outcomes.
\Needspace{0.30\textheight}
\subsection{Probability available at selected decisions}
\begin{figure}[H]
\centering\includegraphics[width=\linewidth]{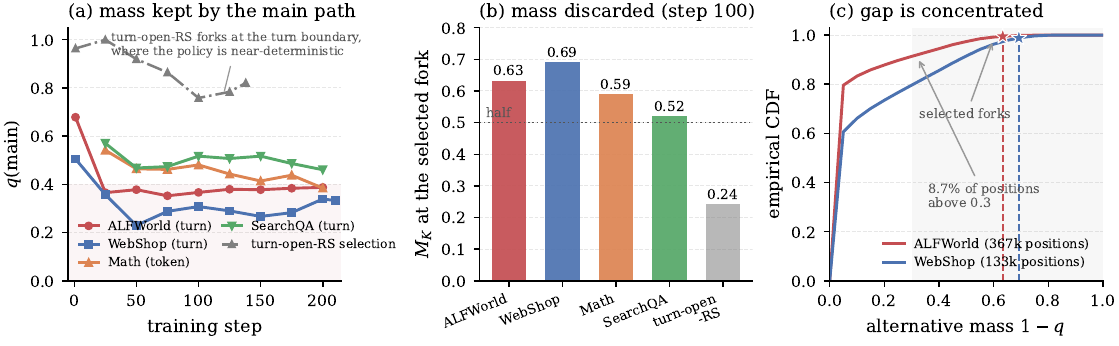}
\caption{Probability mass available for alternative decisions. (a) ``Mass kept by the main path'' is the probability of the sampled main token, $\pi_{\theta_{\rm old}}(a_0\mid s)$. (b) ``Mass discarded'' and $M_K$ denote the alternative-token mass $1-\pi_{\theta_{\rm old}}(a_0\mid s)$, i.e., the mass left unrealized by the main draw. (c) All supervised positions compared with selected forks. ALFWorld and WebShop use training dumps; mathematics and Search-QA use reconstructed summaries. In the plot, $q$ denotes $\pi_{\theta_{\rm old}}$ and turn-open-RS denotes Turn-initial branching.}
\label{fig:altmass}
\end{figure}
\begin{figure}[H]
\centering\includegraphics[width=.78\linewidth]{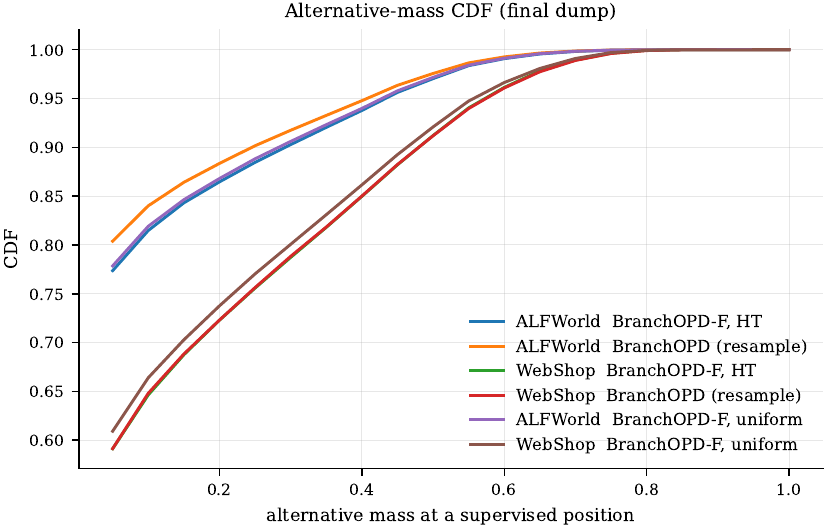}
\caption{Empirical CDF of alternative mass over supervised positions (ALFWorld step 200, WebShop step 210), accumulated over 20 equal-width bins. BranchOPD-F denotes forced-token expansion (Table~\ref{tab:t9}), and ``uniform'' denotes uniform weights on its branches.}
\label{fig:cdf}
\end{figure}

At selected forks, the alternative-token mass $1-\pi_{\theta_{\rm old}}(a_0\mid s)$ is about 0.63 on ALFWorld and 0.69 on WebShop (Figure~\ref{fig:altmass}). Ordinary supervised positions are far more concentrated: 80.4\% and 59.1\% of them have alternative mass below 0.05 (Figure~\ref{fig:cdf}). The selected forks are thus decisions with several plausible continuations.
\FloatBarrier

\Needspace{0.30\textheight}
\subsection{Alternative actions across training}
\begin{table}[H]
\centering\small
\caption{Behavior coverage from 64 main trajectories and 128 branches per checkpoint. Action and observation disagreement and NSR are percentages; new actions is a count per branch.}
\label{tab:t5}
\fitTable{%
\begin{tabular}{lcccc}
\toprule
Task / step & Action (\%) & Observation (\%) & NSR (\%) & New actions \\
\midrule
ALFWorld @100 & 74.2 & 74.2 & 55.5 & 2.44 \\
ALFWorld @200 & 74.2 & 73.4 & 57.8 & 3.02 \\
WebShop @100 & 47.7 & 38.3 & 37.5 & 0.68 \\
WebShop @200 & 52.3 & 40.6 & 32.0 & 0.60 \\
\bottomrule
\end{tabular}}
\end{table}

Action disagreement (EBR) is the fraction of branches whose action sequence differs from that of the main trajectory, and observation disagreement is the analogous fraction for observations. New-action coverage (NSR) is the fraction of branches that contain an action absent from the main trajectory, and mean new actions counts such actions per branch. At step 100, EBR is 74.2\% on ALFWorld and 47.7\% on WebShop, and NSR is 55.5\% and 37.5\% (Table~\ref{tab:t5}).
\begin{figure}[H]
\centering\includegraphics[width=.95\linewidth]{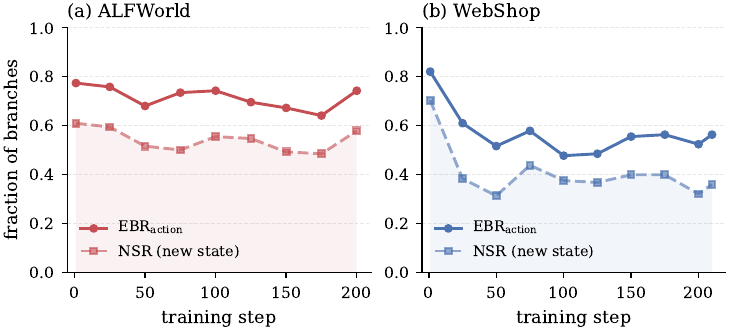}
\caption{Behavior coverage of resampled branches over training. ``New state'' in the plot denotes new-action coverage (NSR). High disagreement and NSR mean that branches reach behavior not present in the main trajectory.}
\label{fig:behavior}
\end{figure}
This coverage persists throughout training (Figure~\ref{fig:behavior}): on WebShop it stabilizes after an initial drop, and on ALFWorld it stays high.
\FloatBarrier

\Needspace{0.30\textheight}
\subsection{Recovery opportunities}
We define the rescue rate $R_\uparrow=\Pr(\text{some branch succeeds}\mid\text{main fails})$ and the harm rate $R_\downarrow=\Pr(\text{all branches fail}\mid\text{main succeeds})$. In the latter case, the successful main trajectory is still used for training.
\begin{table}[H]
\centering\small
\caption{Recovery outcomes (\%). $R_\uparrow$: at least one successful branch given a failed main; $R_\downarrow$: all branches fail given a successful main. The main trajectory is retained.}
\label{tab:t6}
\fitTable{%
\begin{tabular}{lccc}
\toprule
Task / setting & Main success (\%) & $R_\uparrow$ (\%) & $R_\downarrow$ (\%) \\
\midrule
Math B=3,k=2 @200 & 41.4 & 42.7 & 0.0 \\
ALFWorld resample @100 & 48.4 & 21.2 & 19.4 \\
ALFWorld resample @200 & 45.3 & 14.3 & 6.9 \\
WebShop resample @200 & 37.5 & 5.0 & 4.2 \\
Search-QA B=3, pooled dumps & 51.9 & 15.8 & 15.2 \\
\bottomrule
\end{tabular}}
\end{table}

\begin{figure}[H]
\centering\includegraphics[width=\linewidth]{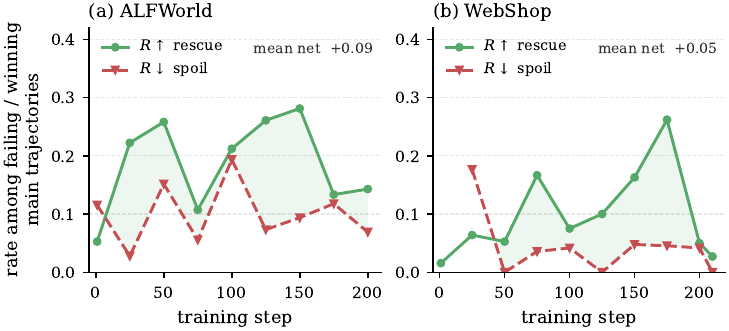}
\caption{Rescue and harm rates during training, each conditioned on the outcome of the main trajectory. ``Mean net'' is $R_\uparrow-R_\downarrow$ averaged over training steps. ``Spoil'' means that all branches fail although the main trajectory succeeds; the main trajectory itself stays in the training batch.}
\label{fig:rescue}
\end{figure}
The rescue rate ranges from 5.0\% on WebShop to 42.7\% on mathematics (Table~\ref{tab:t6}), and successful alternatives keep appearing throughout training (Figure~\ref{fig:rescue}). Such recovered continuations provide corrective supervision and may contribute to the task gains.
\FloatBarrier

\Needspace{0.30\textheight}
\subsection{Where behavior begins to diverge}
\begin{figure}[H]
\centering\includegraphics[width=\linewidth]{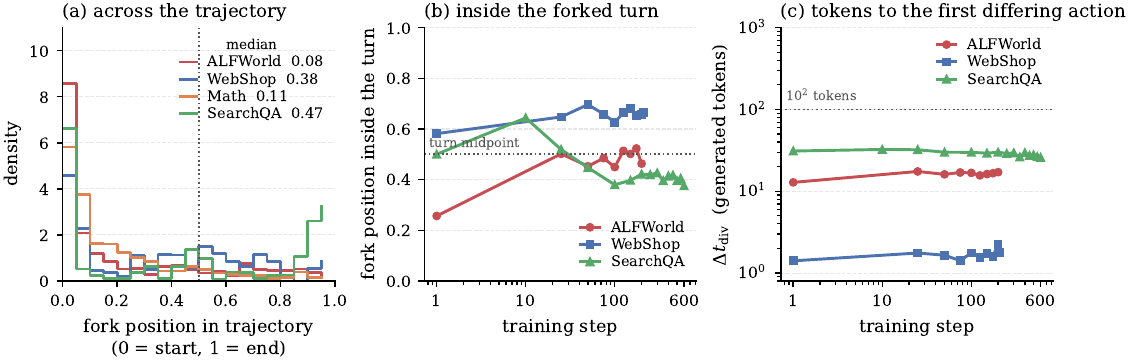}
\caption{Fork location and distance to the first changed action. (a) Fork position relative to the full trajectory and (b) relative to the selected turn, where 0 and 1 mark the start and end. (c) Number of generated student tokens from the fork to the first changed action.}
\label{fig:localization}
\end{figure}
Fork locations depend on the task and often fall inside a turn (Figure~\ref{fig:localization}). The first changed action appears roughly 12--18 generated tokens after the fork on ALFWorld, 1.5--2 on WebShop, and 30 on Search-QA, so a token-level fork can change later actions while reusing the preceding context.
\FloatBarrier

\section{Training and Evaluation Settings}
\label{app:implementation}
\subsection{Shared training configuration}
The default configuration uses $B=3$ forks and $k=2$ continuations per fork. All continuations are kept, including unsuccessful ones, and resampled branches receive no importance weights. Prefix deduplication masks the tokens strictly before the fork and keeps the fork token; environment observations are excluded from the distillation loss. Fork scores mix entropy and teacher--student disagreement with $\alpha=0.5$, using a 128-token disagreement window and the top-20 student log probabilities. In agent tasks, each turn is scored by the mean of its top 20\% eligible-token scores; high-scoring turns are selected with a minimum gap of one turn, and the fork is drawn uniformly from the three highest-scoring eligible tokens of each selected turn. Turn-opening eligibility covers the first 16 tokens of a turn, with no initial-token skip, and is not restricted to turns that follow a tool call. The minimum entropy for eligibility is 0 for agent tasks and 0.5 for mathematics.

Training uses the sampled-token ($k_1$) estimator of the reverse KL, i.e., the detached advantage $\sg[\Delta\log p_t]$, with PPO clipping ratio $\epsilon=0.2$ and token-mean reduction. The teacher is queried for its top-64 log probabilities, and the loss uses the sampled-token log ratio. Recorded loss values are clamped to $[-10,10]$, and log probabilities are floored at $-10$. Task rewards, KL rewards, and a separate actor KL penalty are disabled. Rollouts and branches use temperature 1, top-$p=1$, and no top-$k$ truncation. The learning rate is $10^{-6}$. We train with FSDP, dynamic batching, offloading, gradient checkpointing, and asynchronous vLLM rollout that records log probabilities. For agent tasks, PBR uses an unlikelihood weight $\lambda=0.01$ and a cap of $c_{\rm PBR}=0.02$ on the weighted loss, and $\lambda=0$ for the $-$PBR variants; the forbidden boundary token for Qwen models is 151668.

\begin{table}[ht]
\centering\small
\captionsetup{justification=centering}
\caption{Training settings.}
\label{tab:training_settings}
\fitTable{%
\begin{tabular}{lcccc}
\toprule Setting & ALFWorld & WebShop & Search-QA & Mathematics\\\midrule
Training examples & 6,400 & 6,782 & NR & 17,398\\
Train/PPO batch size & 64/64 & 64/64 & 128/128 & 128/128\\
Prompt limit & 2,048 & 4,096 & 2,048 & 2,048\\
Response limit & 24,576 & 16,384 & 8,192 & 16,384\\
Per-turn tokens & 512 & 512 & 512 & N/A\\
Environment step limit & 50 & 15 & 4 & N/A\\
Environment pool & 16 & 8 & NR & N/A\\
Training duration & 2 epochs & 4 epochs & 600 steps & 2 epochs\\
Rollout/teacher TP & 2/8 & 2/8 & 2/4 & 1/2\\
Rollout/teacher GPUs & 8/8 & 8/8 & 4/4 & 8/8\\
Memory fractions & 0.45/0.75 & 0.45/0.75 & 0.40/0.70 & 0.45/0.65\\
PPO token limit & 32,768 & 24,576 & 16,384 & 49,152\\\bottomrule
\end{tabular}}
\end{table}
Table~\ref{tab:training_settings} lists the per-task settings. Prompt, response, per-turn, and PPO limits are in tokens; NR marks an unreported setting and N/A a setting that does not apply. Mathematics training runs for two epochs over 17,398 examples, i.e., $\lfloor17{,}398/128\rfloor=135$ steps per epoch and 270 steps in total, with checkpointing and evaluation every 50 steps. For mathematics, rollout and teacher inference each run on a separate eight-GPU machine, with tensor-parallel degrees 1 and 2 and GPU memory fractions 0.45 and 0.65.

\subsection{Additional mathematics baselines}
\label{app:math_baselines}
\paragraph{Offline knowledge distillation (KD).}
The teacher samples one response per training problem to form a fixed dataset. The student is trained on these sequences with an equally weighted (0.5 each) sum of token cross-entropy and the forward KL from teacher to student, computed over the teacher's top-16 tokens.

\paragraph{Outcome-based GRPO.}
The student generates eight responses per problem and receives a binary answer-correctness reward; group-relative advantages computed from these rewards drive the GRPO updates. No teacher is used. KD and GRPO are trained without PBR in Table~\ref{tab:math}.

\paragraph{Entropy-aware OPD (EOPD).}
EOPD \citep{ref58} adds a forward-KL term to reverse-KL OPD at tokens where the teacher entropy exceeds a threshold $h_0$:
\begin{equation}
\ell_{\mathrm{EOPD}}(s)=\ell_{\mathrm{OPD}}(s)+\beta\,\mathbf{1}\{H(\pi_T(\cdot\mid s))>h_0\}\KL(\widetilde \pi_T\Vert\pi_\theta),
\end{equation}
where $\ell_{\mathrm{OPD}}$ is the reverse-KL surrogate, $\widetilde \pi_T$ is the teacher distribution renormalized over its top-$k$ tokens, and $\beta$ weights the added term. EOPD changes the local target but samples no additional suffixes. It also differs from Entropy-Weighted OPD, which only reweights the token losses of a single trajectory.

\paragraph{Sparse probing and outcome calibration (SPOT).}
SPOT \citep{ref79} selects positions by normalized teacher entropy, teacher top-$k$ probability mass, and teacher--student mismatch. At each selected position it forces teacher-proposed candidate tokens, lets the student continue, and scores the continuations with an outcome verifier. For positions with a positively valued candidate, it constructs the target
\begin{equation}
p_{\mathrm{cal}}(v\mid s)=\frac{\widetilde \pi_T(v\mid s)\exp(\gamma\widehat V(s,v))}{\sum_{u\in\mathcal K(s)}\widetilde \pi_T(u\mid s)\exp(\gamma\widehat V(s,u))},
\end{equation}
where $\mathcal K(s)$ is the teacher candidate set, $\widehat V$ the verifier-based continuation value, and $\gamma$ the calibration strength. A cross-entropy loss toward this target is added to OPD. \method{} instead resamples from the student and applies dense teacher supervision along the new suffixes. All four additional baselines are trained without PBR in Table~\ref{tab:math}.

\subsection{EOS alignment for the base student and instruction teacher}
The official Qwen3-4B-Base configuration uses EOS token ID 151643 (\texttt{<|endoftext|>}), whereas Qwen3-8B uses ID 151645 (\texttt{<|im\_end|>}) as its primary EOS and accepts both 151645 and 151643 in its generation configuration. In our runs, the student adopts the teacher's EOS configuration and stops on either token (\texttt{eos\_token\_id = [151645, 151643]}). This overrides the base student's single-token default, so teacher and student share the same end-of-response convention for both teacher probabilities and terminal supervision.

\subsection{Evaluation protocol}
For mathematics, we sample 16 answers per problem with temperature 0.6, top-$p=0.95$, and top-$k=20$, with prompt and response limits of 2,048 and 16,384 tokens. For agent tasks, we sample one trajectory per task at temperature 0.4. The model's optional thinking mode is disabled in the generation configuration; protocol tags written in the prompt are handled separately. WebShop uses the small environment with human interaction disabled and seed 42, and holds out 128 of its 6,910 tasks. The WebShop and Search-QA teachers are taken after 125 and 500 GRPO steps, respectively. Search returns three results per call with a 30-second timeout.

ALFWorld SR is computed separately on the 140 ID and 134 OOD tasks, and the ALFWorld aggregate is their unweighted mean. WebShop SR is the fraction of the 128 tasks solved, and Search-QA SR is exact-match accuracy on the same mixed validation set for every model. The agent Avg.\ SR is the unweighted mean of the ALFWorld aggregate, WebShop SR, and Search-QA SR; mathematics Avg is the unweighted mean of the three benchmark accuracies.
\FloatBarrier

\section{Benchmarks and Additional Background}
\label{app:benchmarks}
\subsection{Mathematical reasoning}
MATH contains competition problems in algebra, geometry, number theory, and related subjects~\citep{ref33}. We evaluate on MATH-500, the 500-problem test subset used in \emph{Let's Verify Step by Step}~\citep{ref35}, and on AIME 2024 and 2025. Each AIME year has 30 problems from two 15-problem examinations, with integer answers from 000 to 999. Each problem requires a free-form solution and a final answer. avg@16 averages correctness over 16 independent solutions per problem and then over problems. Training problems come from DAPO~\citep{ref16}.

\subsection{ALFWorld}
ALFWorld~\citep{ref13} is a text-based household environment that grounds language actions in embodied tasks. The agent reads textual observations and issues commands to navigate, inspect, pick up, place, and transform objects, and succeeds if it completes the goal within the step limit. We report success on 140 in-distribution (ID) and 134 out-of-distribution (OOD) tasks, and their unweighted mean. During branch training, the selected interaction state is restored, or the preceding actions are replayed, before generation continues from a token-level fork; observations serve as context and are not supervised.

\subsection{WebShop}
WebShop~\citep{ref14} is a simulated online-shopping benchmark. Given a purchase instruction, the agent searches a catalog, browses product pages, selects options, and buys an item. We report success rate (SR) on 128 held-out tasks.

\subsection{Search-QA}
Retrieval-augmented question answering answers information-seeking questions from retrieved evidence~\citep{ref78}. In our Search-QA setting, a trajectory interleaves search queries, retrieved results, and a final answer. SR is the fraction of questions whose normalized final answer exactly matches an accepted answer. Each search returns up to three results, and at most four interaction steps are allowed. The teacher, the untrained student, and all trained models are evaluated on the same mixed validation set.

\subsection{Benchmarks discussed as background}
GSM8K~\citep{ref32} contains grade-school math word problems, and Minerva~\citep{ref34} studies quantitative reasoning with language models. ScienceWorld~\citep{ref52} tests science-oriented agent reasoning, WebArena~\citep{ref57} evaluates agents in reproducible web environments, and AgentGym~\citep{ref44} covers a range of agent tasks. HotpotQA~\citep{ref53}, TriviaQA~\citep{ref54}, and MuSiQue~\citep{ref55} are question-answering benchmarks, some requiring multi-hop evidence.

\subsection{Additional learning and systems context}
\label{app:background}
Context distillation, self-distillation, and privileged supervision are other ways to transfer information into model parameters~\citep{ref61,ref62,ref63}. Recent work also studies reward extrapolation, skill transfer, hybrid supervision, and teacher retirement~\citep{ref64,ref65,ref66,ref74}. PPO and group-relative optimization are the underlying optimizers~\citep{ref38,ref39,ref40}, and agent training systems such as GiGPO, ArCHer, Agent Lightning, WebRL, and Search-R1 optimize task rewards~\citep{ref41,ref42,ref43,ref45,ref46}, unlike the protocol-event loss used here. Our experiments use the Qwen3 and Qwen2.5 families, and Llama 3 is a related open model family~\citep{ref15,ref50,ref51}. HybridFlow~\citep{ref47}, FlashAttention~\citep{ref48}, and PagedAttention~\citep{ref49} improve execution and memory efficiency, complementing our reduction in generated student tokens. Continual on-policy distillation studies supervision and training distributions that change over time~\citep{ref72}; our analysis instead fixes the rollout policy within each sampling group.
\FloatBarrier

\clearpage
\section{Training Algorithm}
\label{app:algorithm}
\begin{algorithm}[ht]
\caption{Branching on-policy distillation}
\begin{algorithmic}[1]
\Require Student $\pi_\theta$, teacher $\pi_T$, task source, fork budget $B$, continuation budget $k$
\For{each training iteration}
\State Freeze the rollout policy $\pi_{\theta_{\rm old}}\leftarrow\pi_\theta$ for this sampling group
\State Sample a task and generate one main trajectory under $\pi_{\theta_{\rm old}}$
\State Compute eligible-token entropy and teacher--student disagreement scores
\If{mathematical reasoning}
\State Select up to $B$ token-level fork positions
\Else
\State Select turns, then select eligible fork tokens within those turns
\EndIf
\State Initialize training tokens with the main trajectory's generated tokens
\For{each selected fork prefix $s_b$}
\For{$j=1,\ldots,k$}
\State Restore the fork state or replay the prefix actions for this branch, if needed
\State Independently sample a continuation from $\pi_{\theta_{\rm old}}(\cdot\mid s_b)$
\State Keep its generated tokens from the fork onward, whatever its outcome
\EndFor
\EndFor
\State Query teacher log probabilities for retained sampled tokens
\State Exclude observations and duplicate branch-prefix tokens from the loss
\State Form stopped-gradient sampled-token advantages and the clipped token-mean surrogate
\State Construct applicable protocol-boundary events and add the PBR loss
\State Update $\theta$ using the combined objective
\EndFor
\end{algorithmic}
\end{algorithm}
If no eligible fork is found, the iteration trains on the main trajectory alone. All sampled branches contribute regardless of task success. At evaluation, agents generate one trajectory per task, and mathematics uses 16 independent responses per problem.
\FloatBarrier

\clearpage
\section{Prompt Templates}
\label{app:prompts}
The figures below give the prompts used in training and evaluation. Mathematics uses a single-turn template with a boxed final answer. ALFWorld and WebShop separate the initial task instruction from later observations and require structured action outputs. Search-QA additionally specifies the format of retrieval results and the feedback for invalid actions. These templates define the output boundaries that PBR checks.
\begin{figure}[ht]
\centering\includegraphics[width=\linewidth,height=.82\textheight,keepaspectratio]{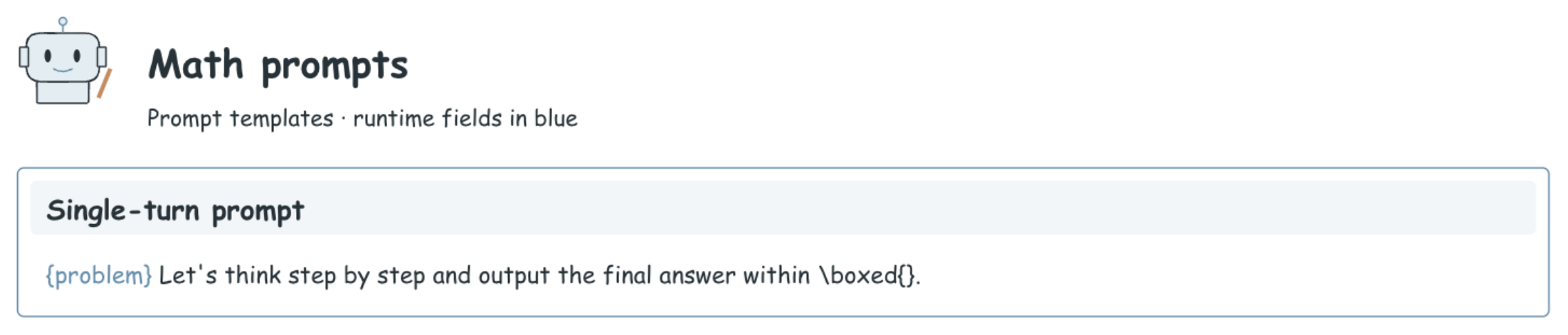}
\caption{Mathematical reasoning prompt templates.}
\end{figure}
\FloatBarrier
\begin{figure}[ht]
\centering\includegraphics[width=\linewidth,height=.86\textheight,keepaspectratio]{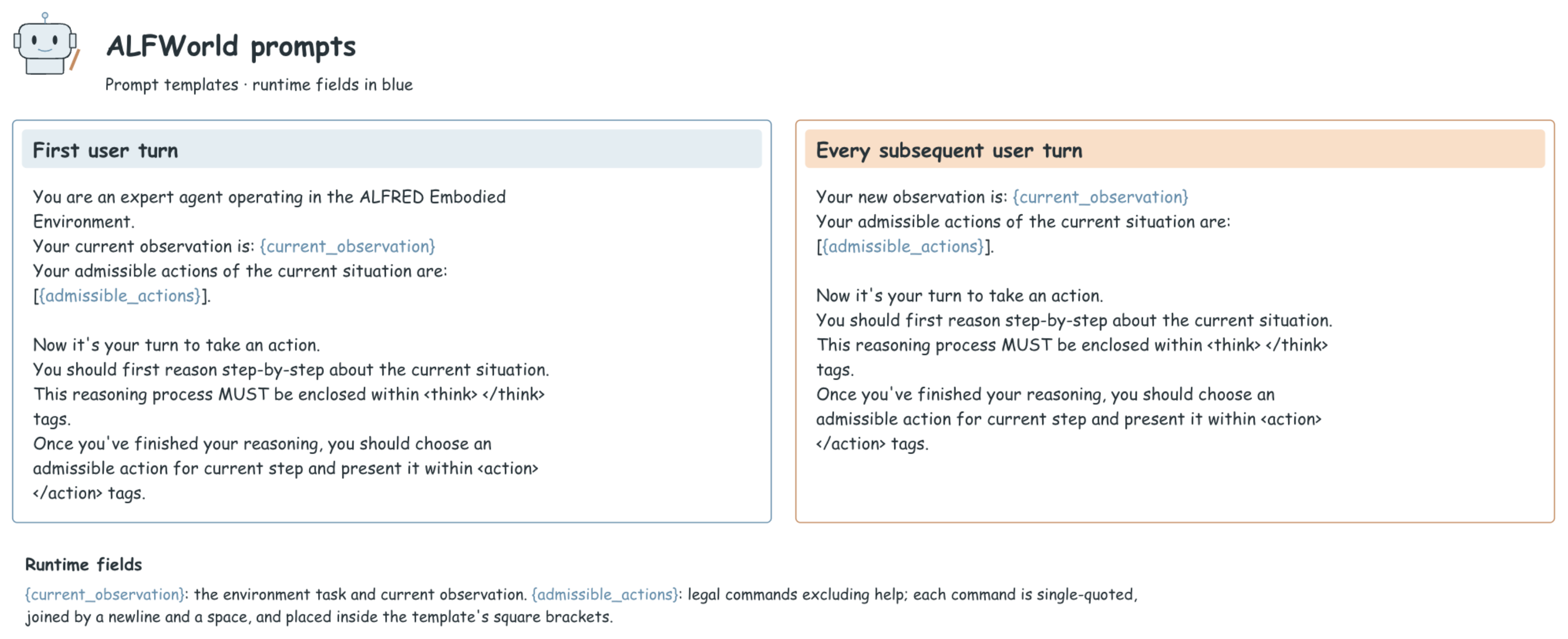}
\caption{ALFWorld prompt templates.}
\end{figure}
\FloatBarrier
\begin{figure}[ht]
\centering\includegraphics[width=\linewidth,height=.86\textheight,keepaspectratio]{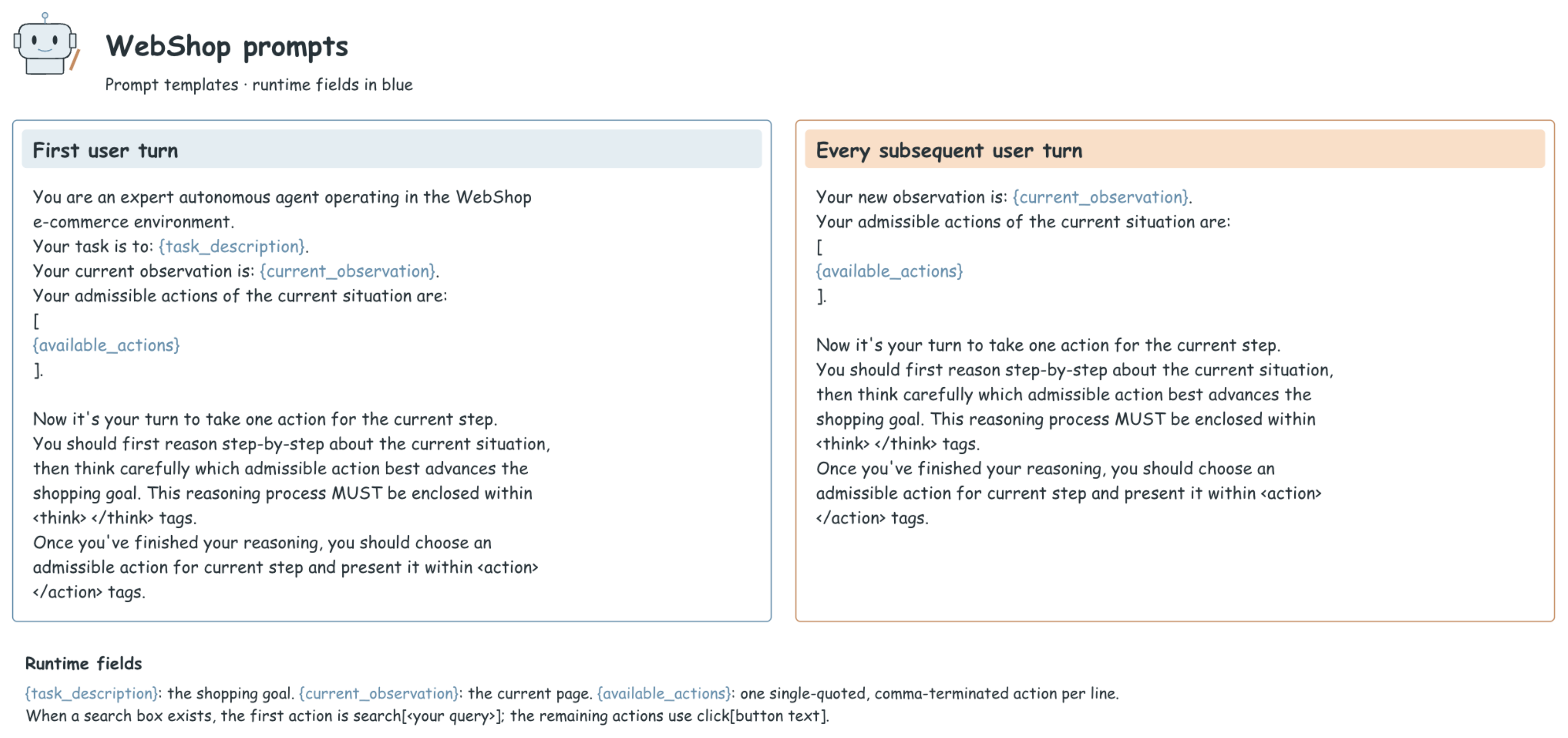}
\caption{WebShop prompt templates.}
\end{figure}
\FloatBarrier
\begin{figure}[ht]
\centering\includegraphics[width=\linewidth,height=.86\textheight,keepaspectratio]{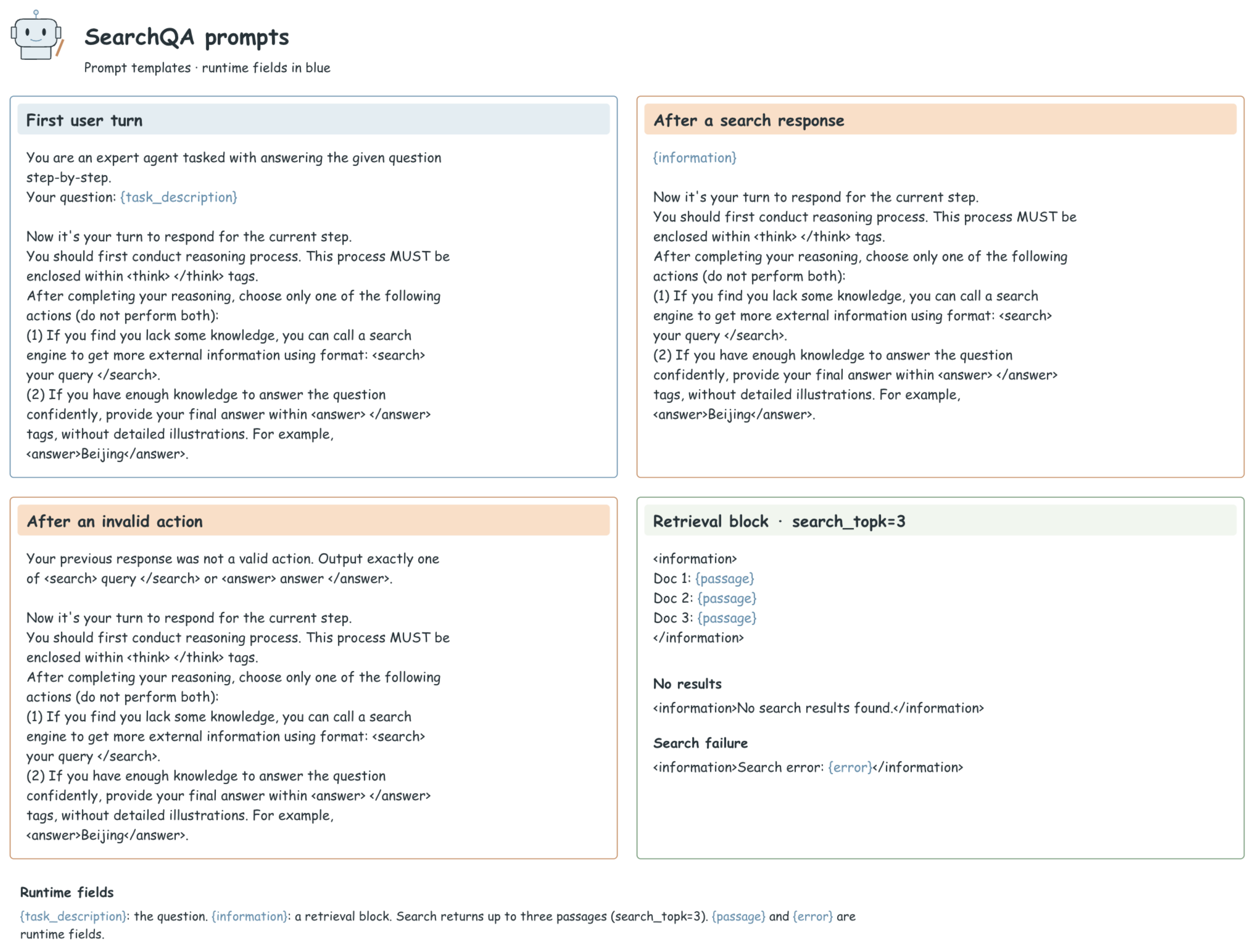}
\caption{Search-QA prompt templates.}
\end{figure}
\FloatBarrier

\end{document}